%% file: acl_latex.tex
\pdfoutput=1
\documentclass[11pt]{article}

\usepackage[preprint]{acl}

\usepackage{times}
\usepackage{latexsym}

\usepackage[T1]{fontenc}
\usepackage[utf8]{inputenc}

\usepackage{microtype}

\usepackage{inconsolata}

\usepackage{graphicx}
\usepackage{hyperref}       % hyperlinks
\usepackage{url}            % simple URL typesetting
\usepackage{booktabs}       % professional-quality tables
\usepackage{amsfonts}       % blackboard math symbols
\usepackage{nicefrac}       % compact symbols for 1/2, etc.
\usepackage{microtype}      % microtypography
\usepackage{xcolor}         % colors
\usepackage{makecell}
\usepackage{amssymb}
\usepackage{graphicx}
\usepackage{amsmath}
\usepackage{multirow}
\usepackage{romannum} % for roman number
\usepackage{algorithm} % for algorithm
\usepackage{algpseudocode} % for algorithm
\usepackage{wrapfig}
\usepackage{ccaption}
\usepackage{setspace}
\usepackage{hhline}

\usepackage{caption}   % For \captionof
\usepackage{subfigure}
\setcitestyle{square}
\usepackage{capt-of}
\usepackage{array}
\newfixedcaption{\outfigcaption}{figure}
\newfixedcaption{\outtabcaption}{table}
\usepackage{authblk}
\usepackage{enumitem}
\usepackage{pifont}
\usepackage{bbding}
\usepackage{fontawesome5}
\usepackage{colortbl}
\usepackage{tabulary}
\usepackage{etoolbox}
\usepackage{cleveref}
\usepackage[most]{tcolorbox} 
\usepackage{longtable}
\usepackage{arydshln} % for dash line in table

\title{What if...?: Thinking Counterfactual Keywords Helps to Mitigate Hallucination in Large Multi-modal Models}

\author{%
Junho Kim \quad\quad
Yeonju Kim \quad\quad
Yong Man Ro \\
Integrated Vision and Language Lab, KAIST \\
\texttt{\{arkimjh, yeonju7.kim, ymro\}@kaist.ac.kr} \\
{\href{https://ivy-lvlm.github.io/Counterfactual-Inception/}{https://ivy-lvlm.github.io/Counterfactual-Inception}}
}

\begin{document}
\pagenumbering{arabic}
\maketitle
\begin{abstract}
    This paper presents a way of enhancing the reliability of Large Multi-modal Models (LMMs) in addressing hallucination, where the models generate cross-modal inconsistent responses. Without additional training, we propose Counterfactual Inception, a novel method that implants counterfactual thinking into LMMs using self-generated counterfactual keywords. Our method is grounded in the concept of counterfactual thinking, a cognitive process where human considers alternative realities, enabling more extensive context exploration. Bridging the human cognition mechanism into LMMs, we aim for the models to engage with and generate responses that span a wider contextual scene understanding, mitigating hallucinatory outputs. We further introduce Plausibility Verification Process (PVP), a simple yet robust keyword constraint that effectively filters out sub-optimal keywords to enable the consistent triggering of counterfactual thinking in the model responses. Comprehensive analyses across various LMMs, including both open-source and proprietary models, corroborate that counterfactual thinking significantly reduces hallucination and helps to broaden contextual understanding based on true visual clues.
\end{abstract}

\section{Introduction}
After witnessing the great success of Large Language Models (LLMs) products, such as ChatGPT~\citep{chatgpt} and Gemini~\citep{Gemini}, the emergence of Large Multi-modal Models (LMMs) naturally followed as the next step towards a unified, general-purpose AI system~\citep{gpt4o, xai2024grokv, reid2024gemini}. In the vision research area, various works~\citep{li2022blip, li2023blip, zhu2023minigpt} have actively resorted LLMs into the vision models due to their remarkable capability of off-the-shelf text generation. Especially when it comes to in-context learning~\citep{brown2020language, alayrac2022flamingo}, prompt engineering~\citep{zhou2022large, bsharat2023principled}, and chain-of-thought~\citep{wei2022chain, kojima2022large, zhang2023multimodal}, vision models can exploit the generation power into the various vision tasks such as visual understanding and reasoning~\citep{yu2022coca, huang2024language}.

% %################################################################################
% Figure
\begin{figure}[t!]
\centering
\includegraphics[width=1.0\linewidth]{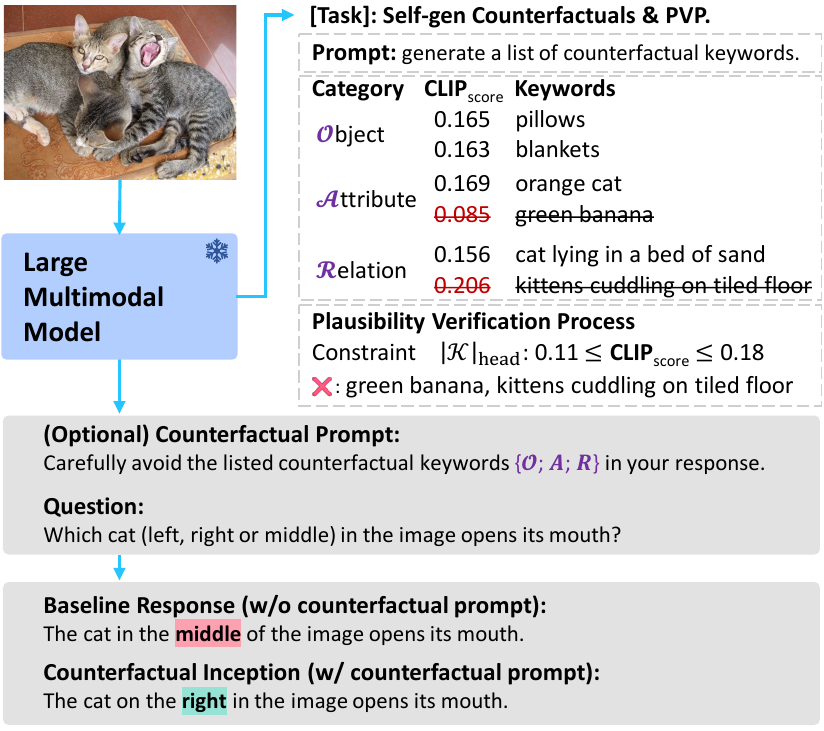}
\vspace*{-0.9cm}
\caption{Counterfactual Inception: LMMs generate counterfactual keywords at the object, attribute, and relation levels, then integrate them with a counterfactual prompt to implant counterfactual thinking to the models. To filter out keywords that are either too similar or too deviated from the visual content, we adopt a robust constraint called PVP.}
\vspace*{-0.5cm}
\label{fig:1}
\end{figure}
% %################################################################################

Although the recent breakthroughs of multi-modal instruction tuning approaches \citep{dai2023instructblip, liu2023visual} unlock enhanced visual proficiency by aligning model responses with human-specific instructions, LMMs still struggle with unexpected hallucination in their responses~\citep{liu2023mitigating, zhou2024analyzing}. The hallucination in LMMs involve false premises, where the models generate incorrect, nonsensical, or unrelated responses for the visual contents. To alleviate the hallucination in LMMs, recent studies have been proposed in the context of curated instruction-tuning~\citep{liu2023mitigating, wang2023vigc}, or integrating visual information using external solvers~\citep{wang2024mitigating, yin2023woodpecker, zhou2024analyzing}. However, they require additional training on the tailored instruction or labor-intensive resources to fine-tune the models~\citep{sun2024aligning, yu2023rlhf}. To step out such limitations and reduce hallucination in a training-free manner, we present a novel way of eliciting an exceptionality capability from LMMs by engaging them to consider alternative counterfactuals.

In our daily life, we ponder \textit{what if...?} scenarios at least once in awhile\textemdash~these sorts of thoughts can be termed as \textit{counterfactual} that is contrary to what actually happened~\citep{menzies2001counterfactual, epstude2008functional}. By thinking of how events might have unfolded differently if we had taken alternative actions (or even seemingly irrelevant thinking), we can enhance cognitive flexibility in the present and identify more about what happens now~\citep{roese1997counterfactual}. Motivated by such human tendency, we delve into the following question: \textit{"Can we elicit counterfactual thinking from LMMs by imagining what-if scenarios and mitigate hallucination in their responses?"}. 

Building on the concept of counterfactuals, we propose Counterfactual Inception, a novel method of implanting counterfactual thinking into LMMs using inconsistent keywords against given visual contents. In our work, we expose LMMs to self-generated counterfactual priors and examine their contextual flexibility in generating responses. Such approach not only allows LMMs to explore a wide range of potential answers but also promotes broader contextual exploration and the consideration of hypothetical narratives. Our findings demonstrate that this thinking enhances the model's ability to engage with and generate responses that spans a wider spectrum of visual understanding, effectively reducing hallucinatory outputs.

Specifically, as illustrated in Fig.~\ref{fig:1}, we instruct LMMs themselves to generate counterfactual keywords at the object-, attribute-, and relation-levels for the visual contents. These keywords are then incorporated into the conditional response generation for user queries with a counterfactual prompt. To consistently promote LMMs to engage in counterfactual thinking, the key challenge is on the optimal selection of counterfactual keywords in triggering the exceptional thought. Accordingly, we present Plausibility Verification Process (PVP), a robust constraint designed to filter out the sub-optimal keywords based on CLIP~\citep{radford2021learning} alignment between the visual contents and their counterfactual keywords. Through extensive analyses on recent LMMs including open-source~\citep{liu2023improved, dong2024internlm, liu2024llavanext, chen2024far} and proprietary models~\citep{Gemini, gpt4v}, we corroborate that Counterfactual Inception helps to alleviate hallucination in general across various benchmarks.

Our contributions can be summarized as follows: (\lowercase\expandafter{\romannumeral1}) we introduce Counterfactual Inception, a novel method that prompts counterfactual thinking into LMMs using deliberately deviated language keywords to mitigate hallucination, (\lowercase\expandafter{\romannumeral2}) we present Plausible Verification Process (PVP), a robust constraint designed to refine the selection of counterfactual keywords, ensuring the optimal trigger of counterfactual thinking in LMMs. (\lowercase\expandafter{\romannumeral3}) Through extensive experiments and analyses on various LMMs, including both open-source and proprietary models, we demonstrate that Counterfactual Inception effectively enhances reliability of model responses across diverse benchmarks.

\section{Related Work}
\label{sec:related} 

\textbf{V+L: Large Multi-modal Models.} 
The release of open-sourced LLMs~\citep{touvron2023llama, vicuna} has spurred active research towards more generalized integration, especially vision-language (VL) modalities. By using the language models as linguistic channels, LMMs can integrate visual information into broader VL understanding tasks~\citep{yang2022unitab, lu2023unifiedio}. After the surge of VL learning~\citep{li2021align, li2022blip, yu2022coca} facilitated cross-modal alignment, recent approach in LMMs is adopting visual instruction-tuning~\citep{dai2023instructblip, liu2023visual,dong2024internlm,chen2024far} on various datasets. LLaVA series~\citep{liu2023visual, liu2023improved, liu2024llavanext} have paved the way for building multi-modality systems that can freely interact with users' instructions. Along with such paradigm, a wide range of advanced architectures and adaptations to specific domains~\citep{lin2023video, li2024llava} have actively explored. Additionally, numerous proprietary LMMs are expanding their capabilities into multi-modal tasks, by releasing advanced products such as Gemini 1.5~\citep{reid2024gemini}, and GPT-4o~\citep{gpt4o}, which allow users to interact with the models through multi-modal channels.

\noindent\textbf{Hallucination in Large Multi-modal Models.} 
Despite the remarkable advancements of LMMs, the major issue of hallucination still persists in their responses. Hallucination refers to the phenomenon where generated texts are inconsistent with the visual contents, one of the long-standing challenges in image captioning~\citep{rohrbach2018object}. When it comes to LMMs, this problem can be worse due to their use of the expressive capabilities of LLMs, which enable more detailed and rich descriptions~\citep{faithscore}. As their representation becomes abundant, the complexity of hallucinations also increases, leading to a multifaceted issue. This includes challenges: (\lowercase\expandafter{\romannumeral1}) the scarcity of large-scale image-text instruction pairs~\citep{liu2023mitigating}, and (\lowercase\expandafter{\romannumeral2}) the entropic gap between visual and textual data~\citep{sun2024aligning}, which can be exacerbated during alignment pre-training.

Recent works have explored various ways to mitigate hallucination, including fine-tuning LMMs with robust instructions~\citep{liu2023mitigating, wang2023vigc}, implementing multi-step LMM-aided reasoning~\citep{wang2024mitigating, yin2023woodpecker, zhou2024analyzing, chen2024unified}, utilizing RLHF~\citep{sun2024aligning, yu2023rlhf} for providing human feedback instructions, and deploying contrastive decoding in the inference phase of LMMs~\citep{leng2023mitigating, woo2024ritual, kim2024code}. More recent hallucination survey compilation can be found in~\citep{liu2024survey, bai2024hallucination}. Our counterfactual method provides a novel approach to reducing hallucinations in LMMs by conditioning them on exceptional thought through counterfactual keywords. We emphasize that our method can achieve significant mitigation of hallucinatory responses without additional fine-tuning or human-resource instructions.

\section{Proposed Method}
\label{sec:method}
\subsection{Counterfactual Keyword Generation}
\label{subsec:key_gen}
Firstly, we can generally categorize the hallucinatory types into three distinct groups\textemdash~nonexistent objects, incorrect object attributes, and erroneous object relations, as found in previous research~\citep{liu2024survey, bai2024hallucination}. To mitigate the hallucination in the model response, our approach is implanting counterfactual thinking into LLMs by harnessing counterfactual keywords. These keywords intentionally do not describe what is visibly apparent but instead provide hypothetical contexts. Importantly, they serve as primary anchors for the contextual exploration for better understanding of true visual clues. Therefore, we concretize counterfactual categories into trinary taxonomy, which can serve plausible alternatives for the visual contents:
\vspace{-4pt}
\begin{itemize}[left=0pt]
    \setlength\itemsep{-0.2em}
    \item \textbf{Object Substitution:} replacing an object in the image with another that could logically occupy the same space but alters the scene's context.    
    \item \textbf{Attribute Modification:} changing an object's color, size, or shape in a way that makes sense visually but leads to a different interpretation.
    \item \textbf{Relational Changes:} adjusting the spatial or interactional relationships between objects to suggest a different narrative within the scene.
\end{itemize}
\vspace{-4pt}

%%%%%%%%%%%%%%%%%%%%%%%%%%%%%%%%%%%%%%%%%%%%%%%%%%%%%%%%%%%%%%%%%%%%%%%%%%%%%%%%%%%
\input{table/inception_example.tex}
%%%%%%%%%%%%%%%%%%%%%%%%%%%%%%%%%%%%%%%%%%%%%%%%%%%%%%%%%%%%%%%%%%%%%%%%%%%%%%%%%%%

Following tailored criteria ($\mathcal{O}$: object, $\mathcal{A}$: attribute, and $\mathcal{R}$: relation), we instruct LMMs themselves to generate three different categorical keywords for the given images, providing plausible but misleading interpretations of the visual contents. Here, obtaining counterfactual keyword is a challenging and complex task for LMMs. Accordingly, we first manually generate a few examples for in-context learning, then design a structured prompt with these seed examples to generate keywords for the categories: $\mathcal{O}{=}{\{o_{i}\}}_{i=0}^{N_{o}}$, $\mathcal{A}{=}{\{a_{i}\}}_{i=0}^{N_{a}}$, and $\mathcal{R}{=}{\{r_{i}\}}_{i=0}^{N_{r}}$, where $N_{o}$, $N_{a}$, and $N_{r}$ represent the different numbers of keywords in each category. We illustrate detailed keyword generation prompts in Table~\ref{table:gen_prompt}. Please see Appendix~\ref{appendix:keygen} for the further explanation of keyword generation.

\subsection{Counterfactual Inception}
After generating the keywords, we implant the counterfactual keywords into LMMs as conditional prior information to guide model responses that disregard these inputs in the generation phase. Specifically, for the given LMMs $M_{\theta}$, parameterized with $\theta$, our objective is generating output sequences $y_{<t+1}{=}[y_{1}, y_{2}, \ldots, y_{t}]$ with given visual content $v$ and textual query $q$. When incorporating self-generated counterfactual keywords to the models, we concatenate all of the keywords generated from a given image into a single list $k{=}[\mathcal{O};\mathcal{A};\mathcal{R}]{\in}\mathbb{R}^{|\mathcal{K}|}$, where $\mathcal{K}$ denotes whole counterfactual keywords set. After that, utilizing these keywords as conditional prior, we can formulate auto-regressive responses of LMMs as follows:
%%%%%%%%%%%%%%%%%%%%%%%%%%%%%%%%%%%%%%%%%%%%%%%%%%%%%%%%%%%%%%%%%%%%%%%%%%%%%%%%%%%
\begin{equation}
\label{eqn:gen}
    p_{\theta}(y {\mid} v, q, k) = \prod_{t=1}^{T}p_{\theta}(y_{t} {\mid} v, q, k, y_{<t}).
\end{equation}
%%%%%%%%%%%%%%%%%%%%%%%%%%%%%%%%%%%%%%%%%%%%%%%%%%%%%%%%%%%%%%%%%%%%%%%%%%%%%%%%%%%
Note that our method can be adapted to existing LMMs in a training-free manner with a specific counterfactual prompt (see Table~\ref{table:cf_prompt}). As exemplified in Table~\ref{table:overview}, we prompt the models to carefully disregard the self-generated counterfactual keywords during their response generation for the user textual query (please see details in~\cref{alg:counterfactual}).

In other words, our method explicitly signal the models to consider alternative explanations anchoring from the self-generated counterfactual keywords. Consequently, our counterfactual approach not only promotes broader contextual understanding but also enhances reliability of the model response. It enables LMMs to focus on true visual clues within the context by incorporating counterfactual information into the response generation, which helps to mitigate hallucination.

\subsection{$\mathcal{K}_{\text{head}}$: Plausibility Verification Process}
Even when we instruct the models to generate keywords, they may not always fulfill our counterfactual intentions\textemdash~for example, even with specific instruction, they might produce completely nonsensical keywords that are irrelevant to the visual content, or generate keywords that are closer to factual rather than counterfactual. Therefore, the key challenge lies in finding the optimal counterfactual keywords $k^{*}{=}\mathcal{K}_{\text{head}}(k)$ that trigger the counterfactual thinking. To analyze the keywords, we randomly sample $500$ images from COCO~\citep{chen2015microsoft} and extract counterfactual keywords from $6$ baselines, totaling 3000 instances and approximately 10K ($\mathcal{O}$), 9.5K ($\mathcal{A}$), and 9.5K ($\mathcal{R}$) keywords in each category, respectively.

\begin{figure}[t!]
\centering
\includegraphics[width=1.0\linewidth]{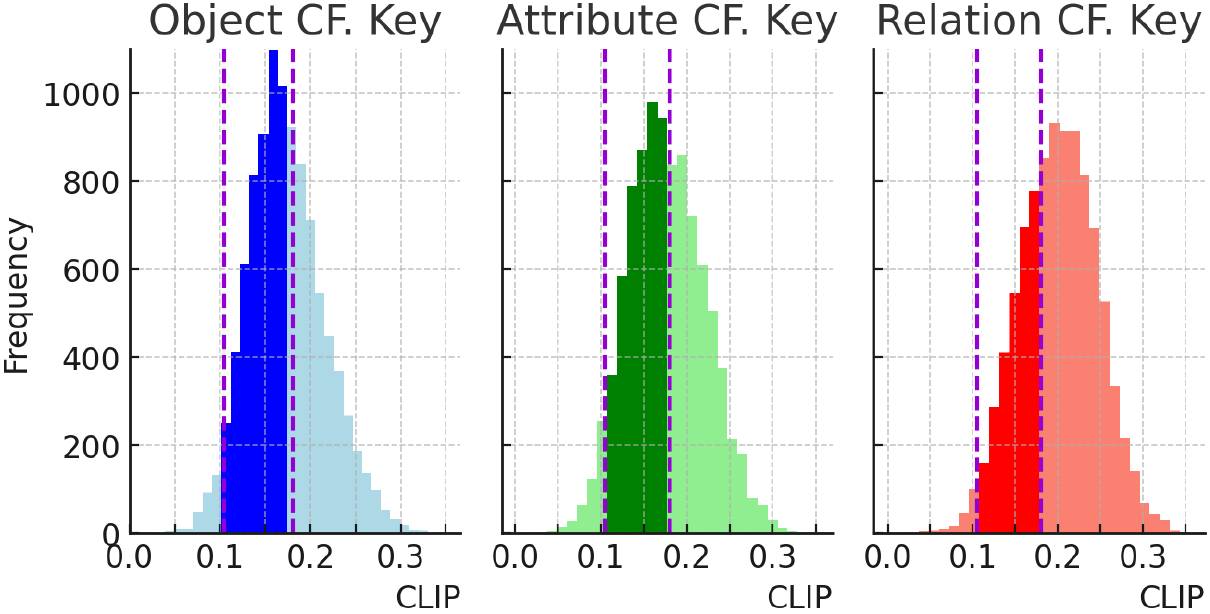}
\vspace*{-1cm}
\caption{Frequency distribution for the counterfactual keywords. The \textcolor{violet}{dashed lines} indicate truncation level. We have empirically observed that the keywords in the upper half of the distribution are closer to factual information rather than counterfactual, thus the lower half, excluding extreme low, is set as the criteria. See Fig.~\ref{fig:6} for the keyword analysis.}
\vspace*{-0.2cm}
\label{fig:2}
\end{figure}

To measure semantic alignment between the counterfactual keywords and visual contents, we employ CLIP~\citep{radford2021learning} and delve into the cross-modal similarity for the text-image pairs. As in Fig.~\ref{fig:2}, the counterfactual keywords, while not directly descriptive, still touch upon concepts or contexts loosely related to the visual contents, leading to a wide range of medium to low scores. Following central limit theorem, the semantic space covered by the keywords has inherent symmetry around a mean value, with fewer keywords being extremely poorly or highly related, creating the bell curve typical of a normal distribution. 

Regarding higher CLIP score suggests a better match\textemdash~that is, the text more accurately or relevantly describes the image, we truncate the counterfactual keyword set based on the score, such that $\mathcal{K}_{\text{head}}(k)=\{k\in\mathcal{K}:\lambda_{\text{bot}} \leq \textbf{CLIP}(v, k) \leq \lambda_{\text{top}}\}$. As in the dashed lines in Fig~\ref{fig:2}, we empirically set the truncation hyperparameter to the lower half of the distribution, but not at the extreme low end, which aligns with the definition of a counterfactual keyword\textemdash~meaningful, yet not direct, alternatives to the visible content. Further analysis in Sec.~\ref{subsec:validity}.

% %################################################################################
% Figure
% %################################################################################

%%%%%%%%%%%%%%%%%%%%%%%%%%%%%%%%%%%%%%%%%%%%%%%%%%%%%%%%%%%%%%%%%%%%%%%%%%%%%%%%%%%
\input{table/discriminative.tex}
%%%%%%%%%%%%%%%%%%%%%%%%%%%%%%%%%%%%%%%%%%%%%%%%%%%%%%%%%%%%%%%%%%%%%%%%%%%%%%%%%%%

\section{Experiments}
\subsection{Experimental Setup}
\label{sec:setup}
\paragraph{Baselines \& Implementation.}
We adopted recent high-performing $6$ LMMs as our baseline models, which can be categorized into open-/closed-source:
(\lowercase\expandafter{\romannumeral1}) open-source: LLaVA-1.5 (13B)~\citep{liu2023improved}, InternLM-XComposer2 (7B)~\citep{dong2024internlm}, LLaVA-NeXT (34B)~\citep{liu2024llavanext}, InternVL 1.5 (26B)~\citep{chen2024far} and (\lowercase\expandafter{\romannumeral2}) proprietary models: Gemini 1.5 Pro~\citep{reid2024gemini} and GPT-4V~\citep{gpt4v} 
%(Please see details in Appendix.~\ref{appendix:baseline}).

For generating counterfactual keyword set $\mathcal{K}$ from each model, we equally used same prompt format in Table~\ref{table:gen_prompt}, but with different guidelines and seed examples. To configure the settings for PVP, CLIP-ViT-L~\citep{radford2021learning} is employed to measure CLIP score (cosine similarity) for the visual contents and the generated counterfactual keyword pairs. We set CLIP score truncation to $0.11$ for lower and $0.18$ for upper boundary. 

\paragraph{Benchmarks and Evaluation Metrics.}
To assess hallucination in LMMs, benchmarks can be sorted into two types: (\romannumeral1) hallucination discrimination, which involves selecting the correct answers from multiple choices, and (\romannumeral2) non-hallucinatory generation, testing the broader range of hallucinations in model responses, measured by either rule-based or GPT-aided methods~\citep{gpt4}. In our experiments, key evaluation benchmarks include POPE~\citep{li-etal-2023-evaluating} and MMVP~\citep{tong2024eyes} for hallucination discrimination, and CHAIR~\citep{rohrbach2018object} and MMHal-Bench~\citep{sun2024aligning} for non-hallucinatory generation (Please see details in Appendix~\ref{appendix:benchmark}): 
\vspace{-7pt}
\begin{itemize}[left=0pt]
    \setlength\itemsep{-0.5em}
    \item \textbf{POPE} uses $9$K image-question pairs from COCO dataset to detect object hallucinations. We exclusively focus on the most challenging, \textit{adversarial} setting. Evaluation metrics are accuracy, precision, recall, and F1-score.    
    \item \textbf{MMVP} measures accuracy for CLIP-blind pairs, which have similar CLIP score but vary visually ($300$ instances \& $9$ visual patterns). Each pattern has curated questions with two response options and scores only if the models identify both pairs.
    \item \textbf{CHAIR} evaluates the proportion of hallucinatory objects in the model responses relative to the total number of objects in the true image caption. It consists of two metric variations: per-sentence and per-instance proportion. 
    \item \textbf{MMHal-Bench} assesses descriptive score and hallucination severity in the model responses using GPT-4 with distinct eight question types. The metric ranges from $0$ to $7$ for the overall score, and the hallucination rate (\%).
\end{itemize}
\vspace{-7pt}

\subsection{Counterfactual Keyword Statistics}
\label{sec:statistics}
\begin{figure}[t!]
\centering
\includegraphics[width=1.0\linewidth]{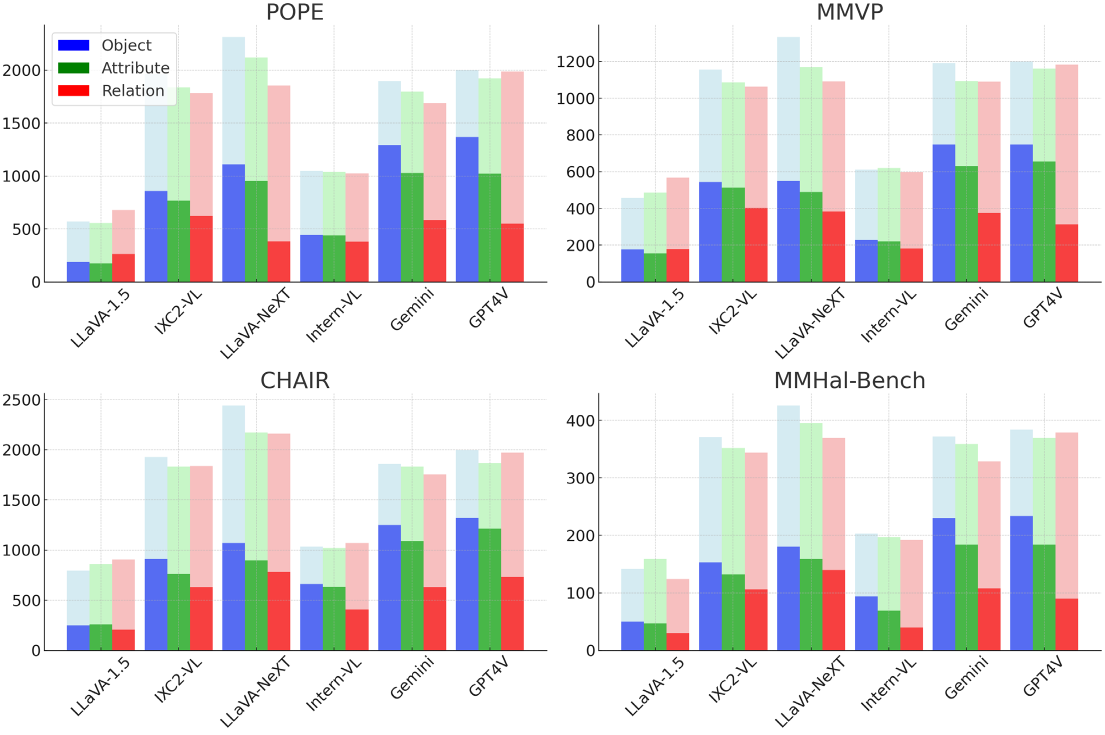}
\vspace*{-1cm}
\caption{The statistical results for the number of counterfactual keywords for $6$ baselines and $4$ benchmarks in each three category. Note that the brighter colors in each bar indicates raw keyword count, and the solid colors are the count after adjusting PVP constraint.}
\vspace*{-0.2cm}
\label{fig:3}
\end{figure}

As in Sec.~\ref{subsec:key_gen}, we first instruct the LMMs themselves to perform the counterfactual keyword generation task and adopt PVP constraint to filter out sub-optimal keywords. For $6$ baselines and $4$ benchmarks we have summarized the keywords statistics in Fig.~\ref{fig:3}. The solid color indicates the frequency after adjusting PVP constraint.

We can observe several interesting findings in the statistics: (\lowercase\expandafter{\romannumeral1}) similar to human perception~\citep{lin2021object}, we can observe LMMs tend to struggle with performing counterfactual thinking in the order of object-, attribute-, and relation-level imagination. This difficulty is clearly shown in the filtered ratios using PVP for each keyword category\textemdash~note that the most filtered category is relation. (\lowercase\expandafter{\romannumeral2}) following the scaling law, the more outperforming models that exploiting larger LLMs shows a better capability of extracting keywords. Especially for proprietary models, they show less than $40\%$ filtered ratio in object- and attribute-level keyword categories, unlike open-source models, which have a filtered ratio of over $50\%$. This results in overall lower average CLIP scores for the keywords generated by both Gemini and GPT-4V compared to the open-sourced models, as in Table~\ref{table:keyword}. More detailed statistics are in Fig.~\ref{fig:keyword} and Appendix~\ref{appendix:statistics}.
% %################################################################################
% Figure
% %################################################################################

\subsection{Experimental Results}
% \vspace*{-0.1cm}
\paragraph{Discriminative Benchmarks.}
The evaluation for discriminative benchmarks is summarized in Table~\ref{table:discriminative}. As in the table, we can observe that overall performance has been improved, compared to the baselines after adopting our methods. Especially, as analyzed in~\citep{liu2023mitigating}, the composition of POPE focuses solely on questioning the existence of objects, rather than their absence (\textit{e.g.,} "Is there \texttt{\{something\}} in the image?"). The combinatorial results of a high accuracy and F1 score indicate that our method can boost the existing LMMs to effectively mitigate hallucination by cautiously confirming \textit{yes} for the existence of objects (\textit{i.e.,} the model does not often \textit{make up} objects). 

We further compare our method with $6$ LMM baselines in MMVP benchmark, which comprehensively assess CLIP-blind pairs for $9$ distinct visual patterns. As shown in the Table~\ref{table:discriminative}, the results indicate significant improvements in average accuracy after adjusting Counterfactual Inception\textemdash~increasing from $5.8\%$ up to $18.53\%$. These improvements show that the counterfactual thinking is indeed helpful to reassess the visual context for the given images without further fine-tuning, leading to reliable responses that capture more relevant facts and complex visual patterns.

%%%%%%%%%%%%%%%%%%%%%%%%%%%%%%%%%%%%%%%%%%%%%%%%%%%%%%%%%%%%%%%%%%%%%%%%%%%%%%%%%%%
\input{table/generative.tex}
%%%%%%%%%%%%%%%%%%%%%%%%%%%%%%%%%%%%%%%%%%%%%%%%%%%%%%%%%%%%%%%%%%%%%%%%%%%%%%%%%%%

\paragraph{Generative Benchmarks.}
Beyond the discriminative benchmarks, which primarily evaluate multiple choice questions, we assess LMM baselines to identify their non-hallucinatory generation capabilities by measuring the proportion of hallucinated contents in their responses. As presented in Table~\ref{table:generative}, our method enhances the overall performance on both CHAIR and MMHal-Bench benchmarks. For CHAIR evaluation, we randomly sample $500$ images from COCO 2014 validation set and prompt ("Please describe this image in detail.") to the models with max generation length of $64$. As in the table, for the both per-sentence ($\text{C}_{\text{S}}$) and per-instance ($\text{C}_{\text{I}}$) results demonstrate consistent improvements in the tasks of long and short description generation across LMM baselines in general.

For the results of MMHal-Bench using GPT-aided evaluation, we clearly observe not only performance gains in the overall score but also a remarkably reduced hallucination ratio. In particular, Gemini 1.5 Pro exhibits a significant hallucination reduction in their responses, with improvements of more than 50\%. From the generative results above, by introducing counterfactuals to LMMs, we demonstrate that our method encourages the model to explore alternative paths, thereby enhancing contextual understanding based on true visual clues and reducing hallucinatory responses.

\input{table/ablation.tex}
%%%%%%%%%%%%%%%%%%%%%%%%%%%%%%%%%%%%%%%%%%%%%%%%%%%%%%%%%%%%%%%%%%%%%%%%%%%%%%%%%%%

\subsection{Analysis on Counterfactual Inception}
\label{sec:ablation}
\paragraph{Ablation Study.} We mainly conduct ablation studies on the following two components: (\lowercase\expandafter{\romannumeral1}) the effectiveness of PVP constraint, which is designed to truncate the self-generated keywords that are either too similar or too deviated and (\lowercase\expandafter{\romannumeral2}) the combinatorial results of using object-, attribute-, and relation-level counterfactual keywords. For the ablation studies, we use two baselines (LLaVA-1.5 and IXC2-VL) along with POPE (discriminative) and mmHal-Bench (generative) benchmarks.

First, as shown in Table~\ref{table:ablation}, the existence of PVP constraint can significantly boost benchmark performances, indicating that the selection of optimal keywords is an important factor for counterfactual thinking. This indicates that disregarding too similar (closer to factual) or too deviated keywords potentially provokes ill-posed response generation and leads to cross-modal inconsistency. Through this ablation, we demonstrate that PVP, which leverages a simple yet effective truncation method based on the alignment score between visual contents and keywords, is a necessary step for integrating counterfactual keywords into LMMs without additional training. Further discussion is in Appendix~\ref{appendix:fail}.

Next, as in Sec.~\ref{subsec:key_gen}, we mainly generate counterfactual keywords at three different levels of granularity\textemdash~object, attribute, or relation. We analyze how the attribute- and relation-level keywords can further enhance performance by using object-level keywords ($\mathcal{O}$) as the primary anchors for conceptualizing counterfactuals. By comparing the results of ${+}\mathcal{O}$ and ${+}\mathcal{O};\mathcal{A};\mathcal{R}$ with PVP constraint adjusted, we recognize that the conjunction of keywords indeed helps to broaden context awareness, which results in performance improvements and mitigates hallucinatory responses.

% %################################################################################
% Figure
\begin{figure}[t!]
\centering
\includegraphics[width=0.90\linewidth]{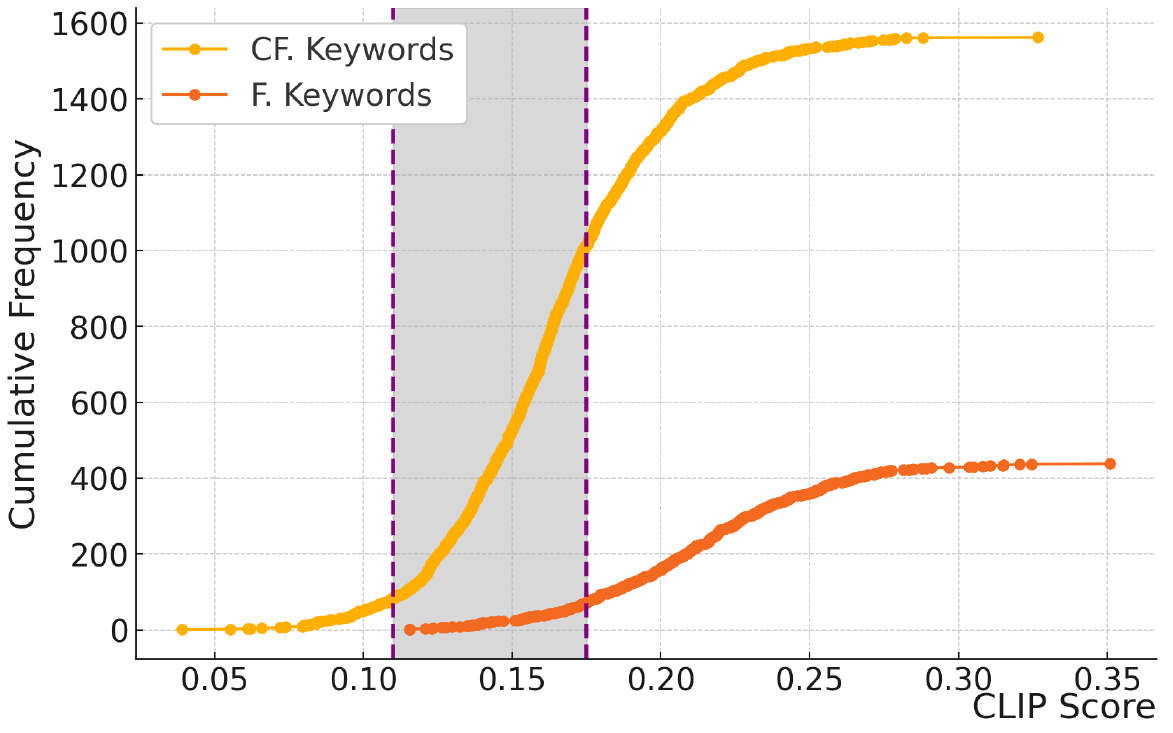}
\vspace*{-0.5cm}
\caption{The cumulative frequency distribution along the scores for COCO dataset with 6 baselines. The \textcolor{violet}{dashed lines} indicates PVP constraint area. }
% \vspace*{-0.2cm}
\label{fig:4}
\end{figure}
% %################################################################################

\paragraph{Validity on Counterfactual Keywords.}
\label{subsec:validity}
We explore the validity of generated counterfactual keywords and the use of PVP constraint by analyzing their distribution across CLIP scores. First, since no ground truth labels for the self-generated keywords, we randomly sampled $100$ images from COCO 2014 validation set and manually determine whether the keywords were closer to counterfactual or factual for the given images (binary task)\textemdash~total $2$K generated keywords integrated from whole $6$ baselines. After that, as illustrated in Fig.~\ref{fig:4}, we visualize the cumulative frequency of each sample based on their CLIP score and analyze distribution with the gray colored PVP constraint area. 

The thresholds of PVP constraint are depicted as purple dashed lines for distinguishing optimal counterfactual keywords. In PVP constraint area, we can observe that a large number of yellow scatter points, categorized as counterfactual keywords, are included in the gray zone with a steep slope. In addition, the orange distribution of factual keywords are mostly located above the upper threshold. In summary, we highlight the robustness of our refinement method in identifying optimal counterfactual keywords. Note that extreme cases (either too similar or too deviated) are sparsely distributed at both extremes and filtered out through PVP constraint.

% %################################################################################
% Figure
\begin{figure}[t!]
\centering
\includegraphics[width=1.0\linewidth]{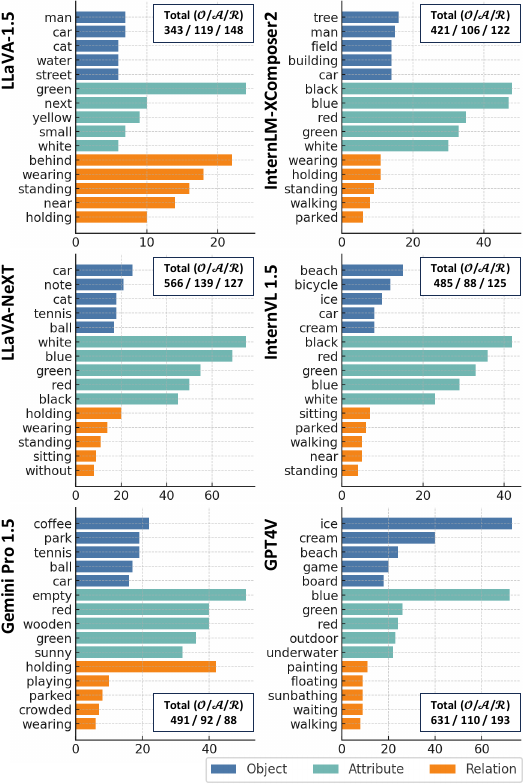}
\vspace*{-0.9cm}
\caption{The graphical results of Top-$5$ words occurrence using morphological analysis (NLTK) in counterfactual keywords. Each legend box indicates total number words in object, attribute, and relation keyword category, respectively.}
\vspace*{-0.4cm}
\label{fig:5}
\end{figure}
% %################################################################################

\begin{figure}[t!]
\centering
\includegraphics[width=1.0\linewidth]{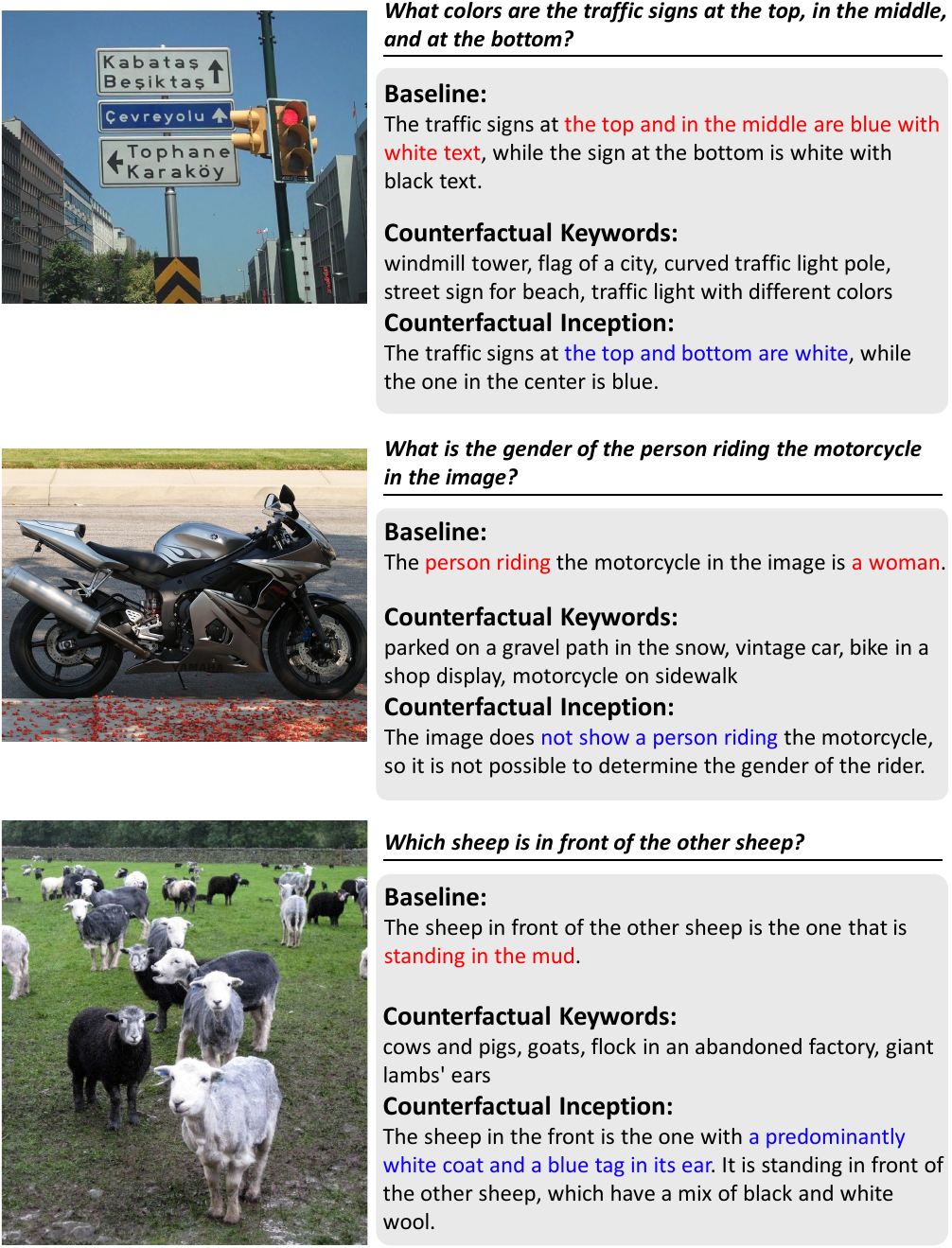}
\vspace*{-0.9cm}
\caption{Case study on MMHal-Bench using the highest-performing model (InternVL 1.5). The hallucinatory responses are marked as 
\textcolor{red}{red}, and the refined responses are \textcolor{blue}{blue} using ours.}
\vspace*{-0.3cm}
\label{fig:6}
\end{figure}

\paragraph{Closer Look at Counterfactual Keywords.}
As an additional analysis, we explore the counterfactual keywords that frequently occurred in each of $6$ baselines for the same $500$ images sampled from COCO 2014, which can reveal word-level distribution and potential bias when generating the keywords. To do that, we tokenize the counterfactual keywords for each category: $\mathcal{O}$, $\mathcal{A}$, and $\mathcal{R}$ with PVP constraint. Then, we conduct a morphological analysis for each category using the following criteria: $\mathcal{O}$ for nouns, $\mathcal{A}$ for adjectives, and $\mathcal{R}$ for adverbs and verbs. In Fig.~\ref{fig:6}, we visualize the top-$5$ morpheme words for each category. As in the figure, we can observe that $\mathcal{A}$ keywords tend to focus on colors when modifying attributes, while both $\mathcal{O}$ and $\mathcal{R}$ are relatively evenly distributed in general, especially considering the low count of top-1 words and total categorical counts. Interestingly, we find that GPT4V shows a notable bias towards "ice" in its generation of counterfactual keywords ($\mathcal{O}$)\textemdash~ice cream, iced tea, iced donuts, etc,. Such bias may the frequently occurred words in its training data, reflecting a specific weakness of the model's ability to generate diverse alternatives. Also this indicates the potential availability of counterfactual keywords as revealing generative vulnerabilities in the alternative responses.

% %################################################################################
% Figure
% %################################################################################

\paragraph{Case Study of Counterfactual Inception.}
The case studies are depicted in Fig.~\ref{fig:6} for the image-question pairs on MMHal-Bench, where it evaluate the degree of hallucination in the generated model responses. As shown in the figure, our method mitigates hallucinatory responses and answers grounded on the true visual clues in the image (not solely based on the biases). We highlight that this is mainly due to the counterfactual keywords\textemdash~plausible but misleading visual interpretations, which expand visual understanding by using these keywords as the primary anchor, thereby enabling broader contextual exploration based on alternative visual contents. We include additional qualitative results and failure cases in Appendix~\ref{appendix:qualitative}.

\section{Conclusion}
In this work, we propose a novel method of reducing hallucination in LMMs, Counterfactual Inception. By integrating counterfactual thinking to the models through self-generated keywords, our approach improves the reliability of model responses. The introduction of Plausibility Verification Process (PVP) further ensures the precision of selecting counterfactual keywords to implant counterfactual thinking. Our extensive analyses across various models and benchmarks corroborate that our approach can effectively trigger exceptional thought to the models without additional training and mitigate hallucination in their responses.

\section{Limitation and Future Scope}
\label{sec:discuss}
Our study introduces Counterfactual Inception, implanting counterfactual thinking into LMMs and demonstrates that conditioning on counterfactual keywords is helpful to mitigate hallucinatory response generation. Despite our new findings, our work reveals several limitations to discuss and future research direction for further exploration. 

First, even if we have examined the recent outperforming baselines with varying model sizes including both open-source and closed-source, due to limited budget and computational power, our work restricted to investigate how the model sizes can affect the capability of implanting counterfactual thinking and the degree of hallucination in their responses. This leaves an open question to figure out the impacts of counterfactual thinking across smaller and larger size of LMMs.

Additionally, while we introduced a simple yet effective PVP constraint to filter out counterfactual keywords, its optimality can be enhanced with a more rigorous filtering mechanism. As we investigated in Sec.~\ref{sec:ablation}, selecting optimal counterfactual keywords significantly affects hallucinatory generation. As discussed in Appendix~\ref{appendix:fail}, incorrectly assigned counterfactual keywords can provoke ill-posed response generation, such as parroting keywords\textemdash~this tendency is exacerbated in smaller models. This suggests a further need to explore more effective methods for identifying optimal counterfactual keywords as a future research direction.

\bibliography{reference}

\clearpage
\newpage

\appendix

\section{Benchmark and Metric}
\label{appendix:benchmark}
We additionally explain the benchmarks details for better understanding of their data statistics and metrics to evaludate hallcination.

\subsection{Discriminative Benchmark}%\mbox{}\\
\textbf{POPE}~\citep{li-etal-2023-evaluating} (Polling-based Object Probing
Evaluation) is designed to detect object hallucinations using $9$K image-question pairs. The questions are about the presence of objects (\textit{e.g.,} "\texttt{Is there a person in the image?"}) and are categorized into three sampling settings based on the selection method of nonexistent objects: \textit{random}, \textit{popular}, and \textit{adversarial}. In the random setting, nonexistent objects are chosen randomly. In the popular setting, objects are selected from a pool of those most frequently occurring, whereas in the adversarial setting, objects that often co-occur but are absent in the image are chosen. In our experiment, we focus exclusively on \textit{adversarial} setting, as it is the most challenging setting than the others and better represents the complex hallucination aspects of real-world adaptation. The evaluation metrics used are accuracy, precision, recall, and F1-score.

\textbf{MMVP}~\citep{tong2024eyes} (Multi-modal Visual Patterns) aims to identify \textit{CLIP-blind pairs} that are considered similar by CLIP but have distinct visual semantics. It contains 150 pairs with 300 questions across 9 visual patterns: Orientation and Direction (\textbf{\faCompass}), Presence of Specific Features (\textbf{\faSearch}), State and Condition (\textbf{\faSync}),  Quantity and Count (\textbf{\faSortNumericUp}), Positional and Relational Context (\textbf{\faMapPin}), Color and Appearance (\textbf{\faPalette}), Structural and Physical Characteristics (\textbf{\faCogs}), Text (\textbf{\faFont}), Viewpoint and Perspective (\textbf{\faCamera}). The questions are carefully designed to ask the details that CLIP vision encoder ignores and provides two options to select (\textit{e.g.,} "\texttt{Where is the yellow animal's head lying in this image? (a)Floor (b)Carpet}). Accuracy is used as the evaluation metric for each of the $9$ visual patterns, and only when the models correctly predict both pairs is the accuracy considered.

\subsection{Generative Benchmark}
\textbf{CHAIR}~\citep{rohrbach2018object} (Caption Hallucination Assessment with Image Relevance) is a benchmark for evaluating image and caption consistency from the language generation. It calculates the degree of word cardinality intersection between the responses generated by the model and the actual image captions. It uses two variations of the metric, per-sentence ($\text{C}_{\text{S}}$) and per-instance ($\text{C}_{\text{I}}$), to evaluate whether the responses include hallucinated objects:
%%%%%%%%%%%%%%%%%%%%%%%%%%%%%%%%%%%%%%%%%%%%%%%%%%%%%%%%%%%%%%%%%%%%%%%%%%%%%%%%%%%
\begin{equation}
\label{eqn:chair}
    \begin{aligned}
    \text{C}_{\text{S}}&=\frac{|\{\text{sentences w/ hallucinatory object}\}|}{|\{\text{all sentences}\}|}, \\
    \text{C}_{\text{I}}&=\frac{|\{\text{hallucinatory objects}\}|}{|\{\text{all objects mentioned}\}|}.
    \end{aligned}
\end{equation}
%%%%%%%%%%%%%%%%%%%%%%%%%%%%%%%%%%%%%%%%%%%%%%%%%%%%%%%%%%%%%%%%%%%%%%%%%%%%%%%%%%%

For CHAIR evaluation, we randomly sampled $500$ images from COCO 2014 validation and generate model responses with the max length of $64$.

\textbf{MMHal-Bench}~\citep{sun2024aligning} focuses on the evaluation of the degree of hallucination, which is different from the previous LMM benchmarks~\citep{liu2023mmbench}, with GPT-4. The question, response, category names of the image content, and human-generated answer are provided as input to GPT-4. Then, GPT-4 measures the severity of hallucination in a range of $0$ to $7$. The higher score denotes less hallucination. The questions can be sorted into $8$ types: object attribute, adversarial object, comparison, counting, spatial relation, environment, holistic description, and others.

%%%%%%%%%%%%%%%%%%%%%%%%%%%%%%%%%%%%%%%%%%%%%%%%%%%%%%%%%%%%%%%%%%%%%%%%%%%%%%%%%%%
\input{table/prompt_keyword.tex}
%%%%%%%%%%%%%%%%%%%%%%%%%%%%%%%%%%%%%%%%%%%%%%%%%%%%%%%%%%%%%%%%%%%%%%%%%%%%%%%%%%%

\section{Details of Counterfactual Inception}
\subsection{Algorithm}
The better understand of full method, we specified the detailed algorithm of Counterfactul Inception in~\cref{alg:counterfactual}.
%%%%%%%%%%%%%%%%%%%%%%%%%%%%%%%%%%%%%%%%%%%%%%%%%%%%%%%%%%%%%%%%%%%%%%%%%%%%%%%%%%%
\input{algo/algorithm.tex}
%%%%%%%%%%%%%%%%%%%%%%%%%%%%%%%%%%%%%%%%%%%%%%%%%%%%%%%%%%%%%%%%%%%%%%%%%%%%%%%%%%%

\subsection{Keyword Generation}
\label{appendix:keygen}
We have utilized counterfactual keywords to implant counterfactual thinking into LMMs. Due to space limits in the main manuscript, the detailed methodology for generating these keywords is elaborated in this section. In Sec.~\ref{subsec:key_gen} of the main manuscript, we categorized counterfactual keywords in three different taxonomy: object substitution $\mathcal{O}$, attribute modification $\mathcal{A}$, and relational changes $\mathcal{R}$. In generating the counterfactual keywords directly from the LMMs, we discovered that a simple instruction such as "Generate counterfactual keywords that mismatch for the given image" cannot fulfill our initial counterfactual intention. This is because the counterfactual thinking requires models to possess complex reasoning capabilities that capture exceptional clues in both visual and linguistic contexts. 

%%%%%%%%%%%%%%%%%%%%%%%%%%%%%%%%%%%%%%%%%%%%%%%%%%%%%%%%%%%%%%%%%%%%%%%%%%%%%%%%%%%
\input{table/prompt_counterfactual.tex}
%%%%%%%%%%%%%%%%%%%%%%%%%%%%%%%%%%%%%%%%%%%%%%%%%%%%%%%%%%%%%%%%%%%%%%%%%%%%%%%%%%%

%%%%%%%%%%%%%%%%%%%%%%%%%%%%%%%%%%%%%%%%%%%%%%%%%%%%%%%%%%%%%%%%%%%%%%%%%%%%%%%%%%%
\input{table/keyword_statistics.tex}
%%%%%%%%%%%%%%%%%%%%%%%%%%%%%%%%%%%%%%%%%%%%%%%%%%%%%%%%%%%%%%%%%%%%%%%%%%%%%%%%%%%

Referring to comprehensive prompt engineering~\citep{bsharat2023principled}, we found that adopting in-context learning is an effective way of generating plausible yet misleading counterfactual keywords for visual content. We hypothesize that this is achievable due to the diverse pre-training on the language models inside LMMs, which includes a wide array of hypothetical and counterfactual scenarios found in various texts such as literature and speculative fiction. 

Accordingly, we first instruct GPT4V~\citep{gpt4v} to generate seed examples that are not grounded in the true visual clues, from the perspectives of three different views\textemdash~object, attribute, and relation. Then, we manually modify the seed examples to meet our counterfactual design. Consequently, as illustrated in Table~\ref{table:gen_prompt}, we introduce a structured prompt to generate counterfactual keywords in three different granularity with selecting options: $\mathcal{O}$, $\mathcal{A}$, and $\mathcal{R}$.  

% %################################################################################
% Figure
\begin{figure*}[t!]
\centering
\includegraphics[width=0.93\linewidth]{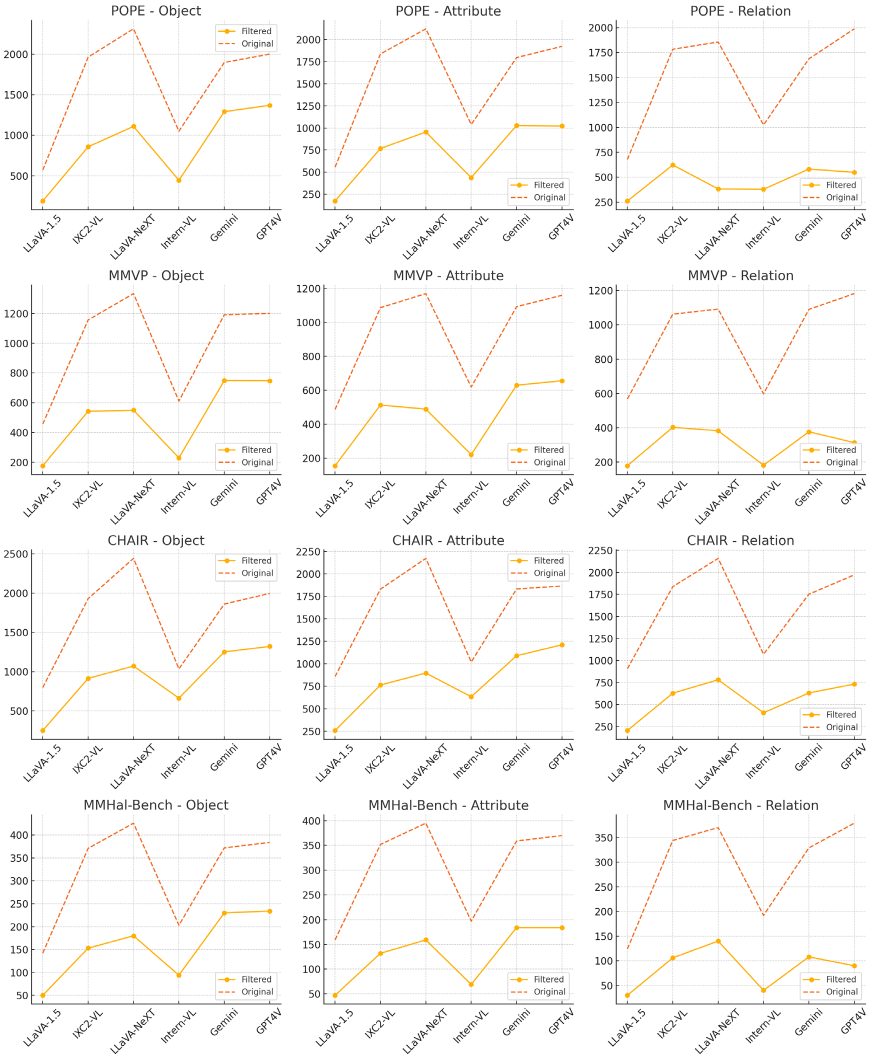}
\vspace*{-0.5cm}
\caption{Detailed analysis on the categorical counterfactual keyword distribution.}
% \vspace*{-0.2cm}
\label{fig:keyword}
\end{figure*}
% %################################################################################

% %################################################################################
% Figure
\begin{figure*}[t!]
\centering
\includegraphics[width=0.9\linewidth]{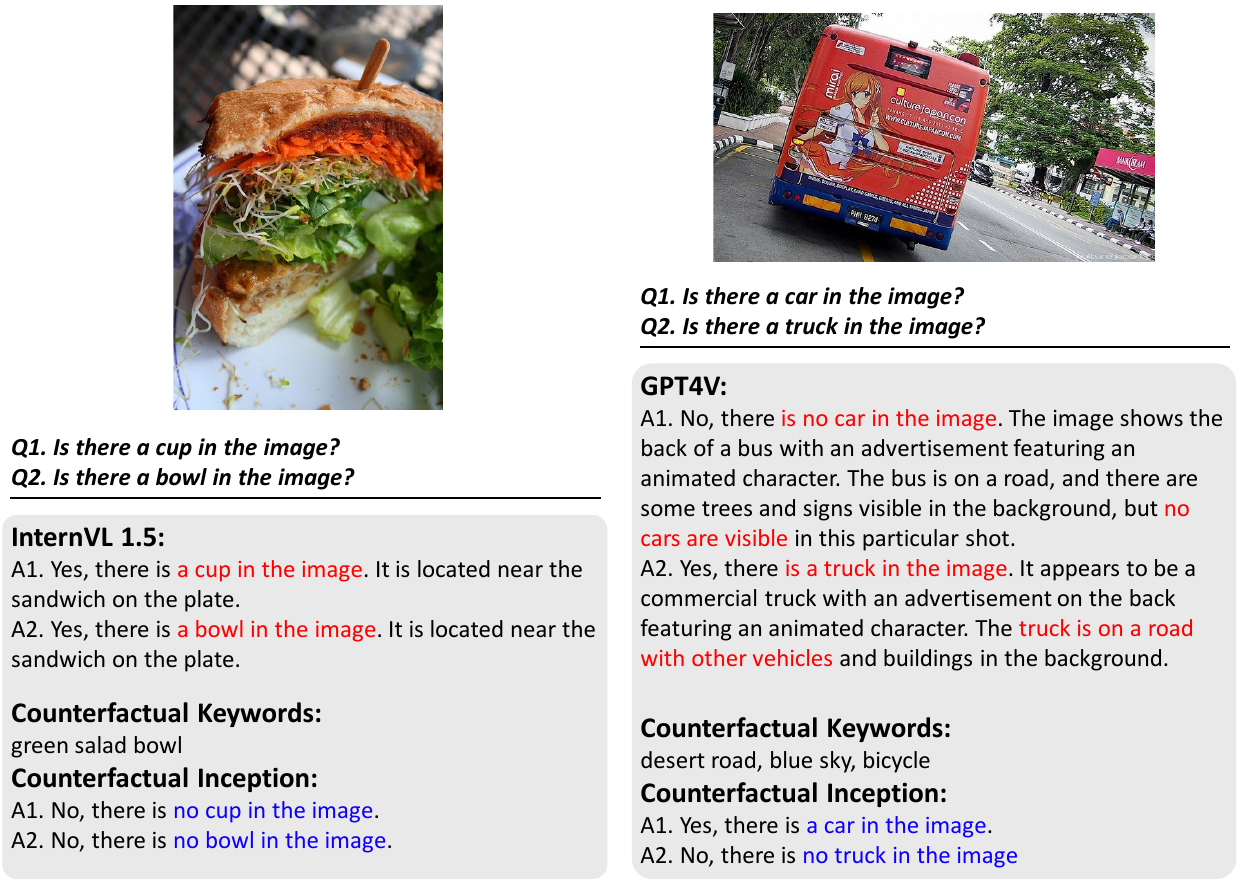}
\vspace*{-0.5cm}
\caption{Additional case study on POPE dataset. The hallucinatory responses are marked as 
\textcolor{red}{red}, and the refined responses are \textcolor{blue}{blue} using ours.}
\vspace*{-0.2cm}
\label{fig:pope}
\end{figure*}
% %################################################################################

\subsection{Counterfactual Prompt}
After obtaining counterfactual keywords, we apply a simple rule-based text pre-processing to filter out non-informative characters such as punctuation marks, stop words, noise words. Subsequently, we designed a specific prompt to integrate the counterfactual keywords with user queries with placeholders, which is then forwarded to the models. As shown in Table~\ref{table:cf_prompt}, we sophisticatedly designed a counterfactual prompt to guide the models in disregarding the extracted counterfactual keywords when generating responses to user queries. We pinpoint that simply implanting the counterfactual prompt with the counterfactual keywords enables the models to mitigate hallucinatory responses without additional training.

\subsection{Details of Keyword Statistics}
\label{appendix:statistics}
In addition to Sec.~\ref{sec:statistics}, we further explore the details of self-generated counterfactual keywords statistics for object, attribute, and relation category. One findings, we can observe as in Table~\ref{table:keyword}, is that the more outperforming LMM baselines show lower average CLIP scores, which indicates better association for the alternatives for the visual clues. Among open-sourced models, we found that InternVL 1.5, which achieved competent performances compared to proprietary multi-modal models, generates relatively a limited number of counterfactual keywords for the given counterfactual instruction. Our assumption of this tendency is on the combined results of its fine-tuning stage, which utilizes text-only data sources such as OpenHermes 2.5~\citep{OpenHermes-2.5}, Alpaca-GPT4~\citep{taori2023alpaca}, ShareGPT~\citep{zheng2024judging}, and COIG-CQIA~\citep{bai2024coig}, and its deeper cross-modal alignment layers, which may leads to focus on the actual clues within the visual context.

\section{Qualitative Assessment}
\label{appendix:qualitative}
\subsection{Additional Case Study}
In our additional case study, we focus on providing further instances demonstrating the effectiveness of our approach, Counterfactual Inception, across various benchmarks. We evaluated our method on discriminative benchmarks such as POPE~\citep{li-etal-2023-evaluating} and MMVP~\citep{tong2024eyes}, generative benchmark MMHal-Bench~\citep{sun2024aligning}.

% %################################################################################
% Figure
\begin{figure}[ht!]
\centering
\includegraphics[width=0.95\linewidth]{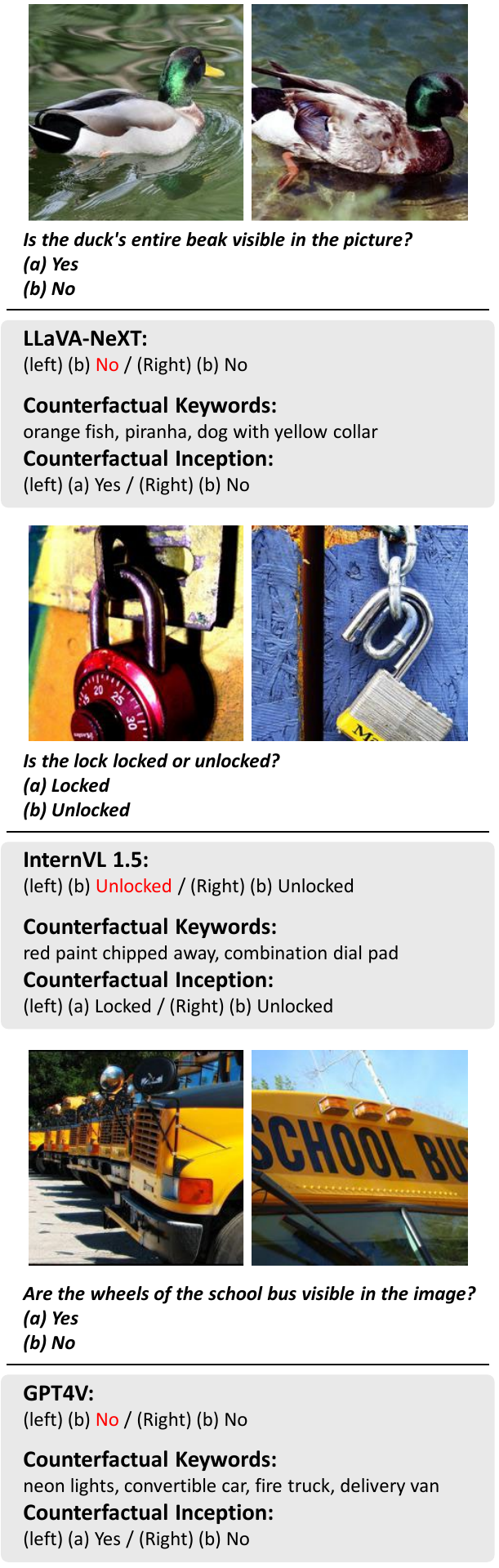}
\vspace*{-0.6cm}
\caption{Additional case study on MMVP dataset. The hallucinatory responses are marked as 
\textcolor{red}{red}, and the refined responses are \textcolor{blue}{blue} using ours.}
\vspace*{-0.2cm}
\label{fig:mmvp}
\end{figure}
% %################################################################################

As in Fig.~\ref{fig:pope} and Fig.~\ref{fig:mmvp}, we illustrate qualitative results for POPE and MMVP datasets, both are discriminative benchmarks where models select answers from the multiple options provided. The utilized models used in this qualitative study are LLaVA-NeXT, InternVL 1.5, and GPT-4V, all of them are the most outperforming multi-modal models in open-source and close-source, respectively. Importantly, we highlight that after conditioning on the given plausible but misleading counterfactual keywords, the baselines demonstrate a better understanding of the true visual clues, enabling a broader contextual exploration that helps to mitigate hallucinatory responses.

\begin{figure*}[t!]
\centering
\includegraphics[width=0.9\linewidth]{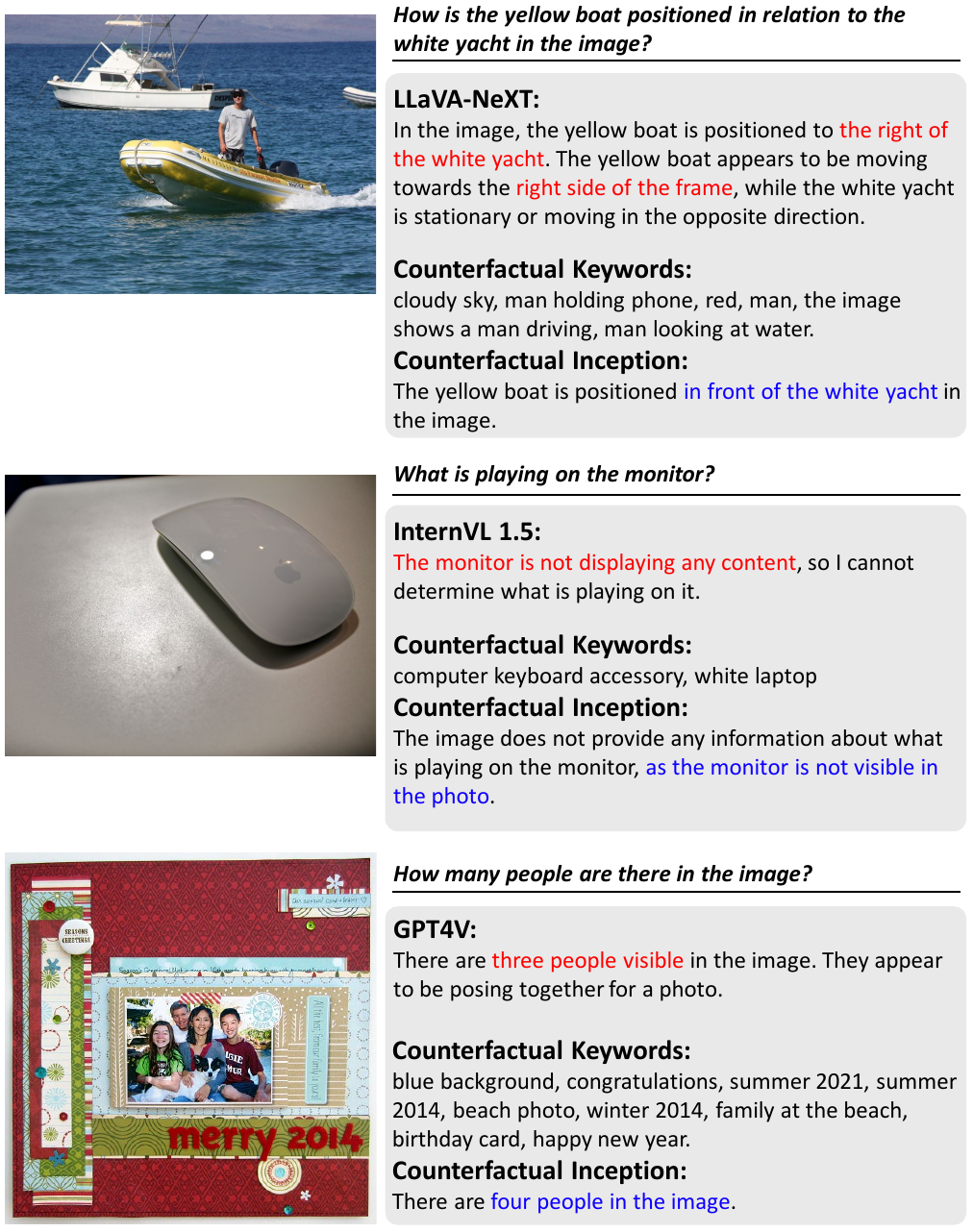}
\vspace*{-0.5cm}
\caption{Additional case study for MMHal-Bench dataset. The hallucinatory responses are marked as 
\textcolor{red}{red}, and the refined responses are \textcolor{blue}{blue} using ours.}
\vspace*{-0.2cm}
\label{fig:mmhal}
\end{figure*}

In Fig.~\ref{fig:mmhal}, we visualize case studies of MMHal-Bench, which is a generative benchmark, to illustrate the effectiveness of Counterfactual Inception in mitigating descriptive hallucination and improving generative ability. The results reveal that the original baselines generate ambiguous or inconsistent responses not grounded on the visual contents, as if the model recognizes non-existent objects. These comprehensive case studies demonstrate that our approach not only enables LMMs to clearly understand the visual context but also significantly enhances their reliability in identifying and describing actual elements present in the visual content, thereby providing more reliable and contextually appropriate responses.

\subsection{Failure Case}
\label{appendix:fail}
\begin{figure}[t!]
\centering
\includegraphics[width=0.81\linewidth]{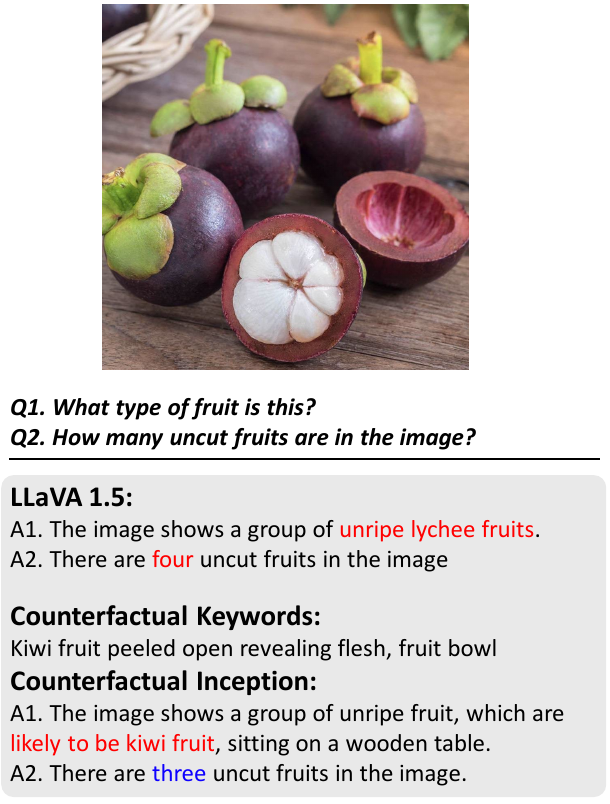}
\vspace*{-0.5cm}
\caption{Failure cases on in-the-wild dataset. The hallucinatory responses are marked as 
\textcolor{red}{red}, and the refined responses are \textcolor{blue}{blue} using ours.}
\vspace*{-0.4cm}
\label{fig:fail}
\end{figure}

Here, we investigate failure cases to understand the limitations of counterfactual thinking as in Fig.~\ref{fig:fail}. Through the analysis, we identified that small models (LLaVA 1.5-13B) sometimes parrots counterfactual keywords in its generated sentences, rather than effectively constructing counterfactual scenarios using these keywords. We hypothesize that this tendency could be linked to the lack of exceptional thought in small models, which potentially leads to the anchoring effect~\citep{tversky1982judgment}, a cognitive bias where initial information disproportionately influences subsequent responses. Although we have proposed a simple and effective PVP constraint to mitigate such negative potential in advance, developing more advanced constraints could be another future research to enhance the counterfactual thinking capabilities of LMMs.

% %################################################################################
% Figure
% %################################################################################

% %################################################################################
% Figure
% %################################################################################

\end{document}

%% file: table/inception_example.tex
%%%%%%%%%%%%%%%%%%%%%%%%%%%%%%%%%%%%%%%%%%%%%%%%%%%%%%%%%%%%%%%%%%%%%%%%%%%%%%%%%%%
\begin{table}[t!]
  \centering
    \scalebox{0.82}{
      \begin{tabular}{p{2.2cm} p{6.2cm} }
        \toprule
        \multicolumn{2}{l}{\bf Example of Counterfactual Inception:}  \\
        \midrule
        &  \includegraphics[height=3cm]{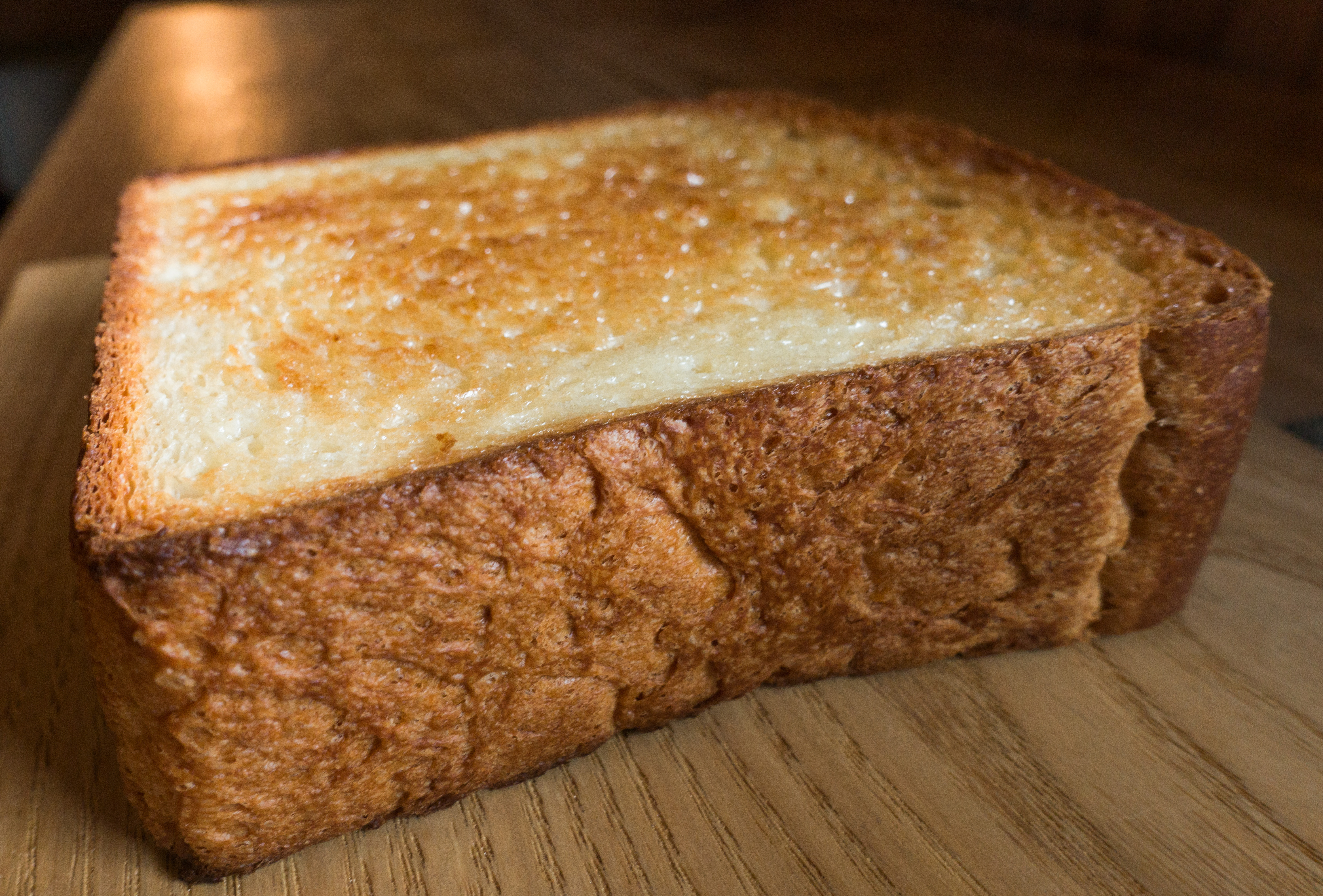} \\ 
        User & Is the jam on the bread made of strawberry? \\
        GT & In fact, this photo only shows a cut and toasted bread slice. There is no jam on it. \\ \midrule 
        Intern-VL 1.5 & Based on the image, \textcolor{red}{the jam on the bread has a deep red color}, which is commonly associated with strawberry jam. However, without a clearer view or additional information, it is not possible to definitively confirm the flavor of the jam. \\ \midrule 
        Counterfactual & \textbf{Counterfactual Prompt:} \\ Inception & Carefully avoid the listed Counterfactual Keywords in your response.\\ \cmidrule{2-2}
        & \textbf{Counterfactual Keywords:} \\ & $\mathcal{O}$: cookie \\ & $\mathcal{A}$: a piece of burnt crusty french baguette \\ & $\mathcal{R}$: toaster on the table \\ \cmidrule{2-2}
        Intern-VL 1.5 & The image \textcolor{blue}{does not show any jam on the bread}, so I cannot confirm if it is made of strawberry or any other fruit.\\
        \bottomrule
      \end{tabular}
    }
    \vspace{-0.4cm}
    \caption{Example of Counterfactual Inception using a baseline Intern-VL 1.5~\citep{chen2024far}.}
    % \vspace{-0.3cm}
    \label{table:overview}  
\end{table}
% %%%%%%%%%%%%%%%%%%%%%%%%%%%%%%%%%%%%%%%%%%%%%%%%%%%%%%%%%%%%%%%%%%%%%%%%%%%%%%%%%%%

%% file: table/discriminative.tex
\begin{table*}[t!]
    \centering
    \renewcommand{\tabcolsep}{2.0mm}
    \resizebox{1.0\linewidth}{!}{
    \begin{tabular}{lccccccccccccccc}
    \Xhline{3\arrayrulewidth}\rule{0pt}{9pt}
    &  & \multicolumn{4}{c}{POPE} & \multicolumn{10}{c}{MMVP} \\ \cmidrule(lr){3-6} \cmidrule(lr){7-16}
    \multirow{-2}{*}{Model} & \multirow{-2}{*}{\#param} & Acc ($\uparrow$) & Prec  & Rec   & F1 ($\uparrow$) & \textbf{\faCompass} & \textbf{\faSearch} & \textbf{\faSync} & \textbf{\faSortNumericUp} & \textbf{\faMapPin} & \textbf{\faPalette} & \textbf{\faCogs} & \textbf{\faFont} & \textbf{\faCamera}      & Avg ($\uparrow$)      \\ \hline
    \multicolumn{16}{l}{\cellcolor[HTML]{EFEFEF}Open-source Models}                                                                                                                                                  \\
    LLaVA-1.5                                  &                           & 84.07          & 90.88 & 75.73 & 82.62          & 22.2 & 50.0 & 23.1 & 20.0 & 40.0 & 60.0 & 36.4 & 37.5 & 16.7 & 35.33          \\
    + Ours                                      & \multirow{-2}{*}{13B}     & \cellcolor[HTML]{DAE8FC} \textbf{85.03} & 93.61 & 75.20 & \cellcolor[HTML]{DAE8FC} \textbf{83.40} & 22.2 & 50.0 & 30.1 & 10.0 & 60.0 & 70.0 & 40.1 & 25.0 & 16.7 & \cellcolor[HTML]{DAE8FC} \textbf{39.33} \\ \hdashline[0.5pt/5pt]
    IXC2-VL                                     &                           & 84.13          & 83.12 & 85.67 & 84.37          & 11.1 & 53.3 & 30.8 & 50.0 & 35.0 & 60.0 & 27.3 & 37.5 & 16.7 & 36.00          \\
    + Ours                                      & \multirow{-2}{*}{7B}      & \cellcolor[HTML]{DAE8FC} \textbf{87.50} & 94.61 & 79.53 & \cellcolor[HTML]{DAE8FC} \textbf{86.42} & 22.2 & 60.0 & 42.3 & 40.0 & 25.0 & 70.0 & 36.4 & 50.0 & 50.0 & \cellcolor[HTML]{DAE8FC} \textbf{42.67} \\ \hdashline[0.5pt/5pt]
    LLaVA-NeXT                                  &                           & \cellcolor[HTML]{DAE8FC} \textbf{86.50}          & 83.86 & 90.40 & 87.01          & 16.7 & 60.0 & 38.5 & 30.0 & 35.0 & 80.0 & 40.9 & 37.5 & 0.0  & 40.67          \\
    + Ours                                      & \multirow{-2}{*}{34B}     &  85.63           & 79.35  & 96.33  & \cellcolor[HTML]{DAE8FC} \textbf{87.02}          & 33.3 & 63.3 & 46.2 & 40.0 & 45.0 & 60.0 & 40.9 & 25.0 & 0.0  & \cellcolor[HTML]{DAE8FC} \textbf{44.67} \\ \hdashline[0.5pt/5pt]
    InternVL 1.5                                &                           & 85.83          & 82.83 & 90.40 & 86.45          & 27.8 & 76.7 & 46.2 & 30.0 & 45.0 & 80.0 & 36.4 & 25.0 & 33.3 & 48.00          \\
    + Ours                                      & \multirow{-2}{*}{26B}     & \cellcolor[HTML]{DAE8FC} \textbf{89.50} & 92.11 & 86.40 & \cellcolor[HTML]{DAE8FC} \textbf{89.16} & 33.3 & 73.3 & 61.5 & 40.0 & 50.0 & 60.0 & 36.4 & 25.0 & 50.0 & \cellcolor[HTML]{DAE8FC} \textbf{51.33} \\ \hline
    \multicolumn{16}{l}{\cellcolor[HTML]{EFEFEF}Proprietary Models}                                                                                                                                           \\
    Gemini 1.5 Pro                              &                           & 80.70          & 85.78 & 73.60 & 79.22          & 27.8 & 53.3 & 38.5 & 40.0 & 55.0 & 40.0 & 45.5 & 62.5 & 66.7 & 46.00          \\
    + Ours                                      & \multirow{-2}{*}{N/A}     & \cellcolor[HTML]{DAE8FC} \textbf{84.09}           & 77.78  & 95.45  & \cellcolor[HTML]{DAE8FC} \textbf{85.71}           & 55.6 & 56.7 & 34.6 & 40.0 & 45.0 & 50.0 & 50.0 & 50.0 & 66.7 & \cellcolor[HTML]{DAE8FC} \textbf{48.67} \\ \hdashline[0.5pt/5pt]
    GPT-4V                                      &                           & 82.70          & 85.50 & 78.80 & 82.00          & 38.9 & 50.0 & 38.5 & 40.0 & 30.0 & 70.0 & 36.4 & 62.5 & 66.7 & 44.00          \\
    + Ours                                      & \multirow{-2}{*}{N/A}     & \cellcolor[HTML]{DAE8FC} \textbf{85.50} & 87.60 & 82.60 & \cellcolor[HTML]{DAE8FC} \textbf{85.07} & 50.0 & 45.5 & 50.0 & 37.5 & 50.0 & 53.3 & 66.7 & 80.0 & 25.0 & \cellcolor[HTML]{DAE8FC} \textbf{48.67} \\
    \Xhline{3\arrayrulewidth}\rule{0pt}{9pt}
    \end{tabular}
    }
    \vspace{-7mm}
    \caption{Evaluation results on discriminative benchmarks. We focus on the most challenging category \textit{adversarial} for POPE~\cite{li-etal-2023-evaluating}. The each column symbol in MMVP~\cite{tong2024eyes} indicates $9$ different visual patterns. We refer Appendix.~\ref{appendix:benchmark} for subset details.}
\label{table:discriminative}
\end{table*}

%% file: table/generative.tex
\begin{table}[t!]
    \centering
    \renewcommand{\tabcolsep}{2.0mm}
    \resizebox{1.0\linewidth}{!}{    
    \begin{tabular}{lccccc}
    \Xhline{3\arrayrulewidth}\rule{0pt}{9pt}
        &                           & \multicolumn{2}{c}{CHAIR} & \multicolumn{2}{c}{MMHal-Bench} \\ \cmidrule(lr){3-4} \cmidrule(lr){5-6}
        \multirow{-2}{*}{Model} & \multirow{-2}{*}{\#param} & $\text{C}_{\text{S}}$ ($\downarrow$)& $\text{C}_{\text{I}}$ ($\downarrow$) & All ($\uparrow$)& Hal ($\downarrow$)\\ \hline
        \multicolumn{6}{l}{\cellcolor[HTML]{EFEFEF}Open-source Models}                                                                               \\
        LLaVA-1.5                                  &                           & 26.4        & 11.12       & 2.39           & 52.1           \\
        + Ours                                      & \multirow{-2}{*}{13B}     & \cellcolor[HTML]{DAE8FC} \textbf{22.4}        & \cellcolor[HTML]{DAE8FC} \textbf{10.94}       & \cellcolor[HTML]{DAE8FC} \textbf{2.54}           & \cellcolor[HTML]{DAE8FC} \textbf{42.7}           \\ \hdashline[0.5pt/5pt]
        IXC2-VL                                     &                           & 24.4        & 9.75        & 3.17           & 29.2           \\
        + Ours                                      & \multirow{-2}{*}{7B}      & \cellcolor[HTML]{DAE8FC} \textbf{20.2}        & \cellcolor[HTML]{DAE8FC} \textbf{8.30}        & \cellcolor[HTML]{DAE8FC} \textbf{3.38}           & \cellcolor[HTML]{DAE8FC} \textbf{25.0}           \\ \hdashline[0.5pt/5pt]
        LLaVA-NeXT                                  &                           & 19.6        & 10.10       & 3.30           & 34.0           \\
        + Ours                                      & \multirow{-2}{*}{34B}     & \cellcolor[HTML]{DAE8FC} \textbf{16.6}        & \cellcolor[HTML]{DAE8FC} \textbf{7.81}        & \cellcolor[HTML]{DAE8FC} \textbf{3.42}           & \cellcolor[HTML]{DAE8FC} \textbf{32.0}           \\ \hdashline[0.5pt/5pt]
        InternVL 1.5                                &                           & 18.2        & 9.00        & 3.15           & 33.3           \\
        + Ours                                      & \multirow{-2}{*}{26B}     & \cellcolor[HTML]{DAE8FC} \textbf{17.8}        & \cellcolor[HTML]{DAE8FC} \textbf{7.93}        & \cellcolor[HTML]{DAE8FC} \textbf{3.42}           & \cellcolor[HTML]{DAE8FC} \textbf{26.0}           \\ \hline
        \multicolumn{6}{l}{\cellcolor[HTML]{EFEFEF}Proprietary Models}                                                                        \\
        Gemini 1.5 Pro                              &                           & 23.4        & \cellcolor[HTML]{DAE8FC} \textbf{12.01}       & 3.62           & 31.0           \\
        + Ours                                      & \multirow{-2}{*}{N/A}     & \cellcolor[HTML]{DAE8FC} \textbf{22.4}        & 12.76       & \cellcolor[HTML]{DAE8FC} \textbf{4.30}           & \cellcolor[HTML]{DAE8FC} \textbf{13.5}           \\ \hdashline[0.5pt/5pt]
        GPT-4V                                      &                           & 20.0        & 9.23        & 3.44           & 28.1           \\
        + Ours                                      & \multirow{-2}{*}{N/A}     & \cellcolor[HTML]{DAE8FC} \textbf{17.8}        & \cellcolor[HTML]{DAE8FC} \textbf{8.67}        & \cellcolor[HTML]{DAE8FC} \textbf{3.47}           & \cellcolor[HTML]{DAE8FC} \textbf{20.8}           \\
    \Xhline{3\arrayrulewidth}\rule{0pt}{9pt}
    \end{tabular}
    }
    \vspace{-8mm}
    \caption{The evaluation results on generative benchmarks. $\text{C}_{\text{S}}$ and $\text{C}_{\text{I}}$ indicates CHAIR metric for sentence- and instance-level, respectively. In MMHal-Bench, "All" indicates overall scores evaluated by GPT-4 and "Hal" denotes the hallucination rate (\%) in the model responses.}
    \vspace{-4mm}
    \label{table:generative}
\end{table}

%% file: table/ablation.tex
\begin{table}[t!]
    \centering
    \renewcommand{\tabcolsep}{2.0mm}
    \resizebox{1.0\linewidth}{!}{    
    \begin{tabular}{llccccc}
    \Xhline{3\arrayrulewidth}\rule{0pt}{9pt}
    \multirow{2}{*}{}         & \multirow{2}{*}{Models} & \multirow{2}{*}{PVP} & \multicolumn{2}{c}{POPE (dis)} & \multicolumn{2}{c}{MMHal-B (gen)} \\ \cmidrule(lr){4-5} \cmidrule(lr){6-7} 
                              &                        &                      & Acc ($\uparrow$) & F1 ($\uparrow$) & All ($\uparrow$) & Hal ($\downarrow$) \\ \midrule
    \rowcolor[HTML]{EFEFEF} 
    \cellcolor[HTML]{EFEFEF}  & LLaVA-1.5          & \cellcolor[HTML]{EFEFEF} & 84.07       & 82.62      & 2.39           & 52.08          \\
    \rowcolor[HTML]{EFEFEF} 
    \multirow{-2}{*}{\cellcolor[HTML]{EFEFEF}Baseline} & IXC2-VL                 & \multirow{-2}{*}{\cellcolor[HTML]{EFEFEF}-} & 84.13       & 84.37      & 3.17           & 29.17          \\  \hdashline[0.5pt/5pt]   
    \multirow{2}{*}{+ $\mathcal{O}$}        & LLaVA-1.5              & \multirow{2}{*}{\ding{55} }  & 83.47       & 81.37      & 2.41             & 46.88              \\
                              & IXC2-VL                &                      & 84.57       & 83.39      & 2.93           & 30.00          \\ \hdashline[0.5pt/5pt]  
    \multirow{2}{*}{+ $\mathcal{O}$}        & LLaVA-1.5              & \multirow{2}{*}{\Checkmark }  & 84.43       & 82.70      & 2.48           & 45.00          \\
                              & IXC2-VL                &                      & 86.53       & 85.29      & 3.21           & 27.00          \\ \hdashline[0.5pt/5pt]  
    \multirow{2}{*}{+ $\mathcal{O}$;$\mathcal{A}$;$\mathcal{R}$}      & LLaVA-1.5              & \multirow{2}{*}{\ding{55} }  & 83.57       & 81.64      & 2.42           & 46.00          \\
                              & IXC2-VL                &                      & 86.13       & 84.89      & 2.79           & 36.46          \\ \hdashline[0.5pt/5pt]  
    \multirow{2}{*}{+ $\mathcal{O}$;$\mathcal{A}$;$\mathcal{R}$}      & LLaVA-1.5              & \multirow{2}{*}{\Checkmark }  & 85.03       & 83.40      & 2.54           & 42.71          \\
                              & IXC2-VL                &                      & 87.50       & 86.42      & 3.38           & 25.00          \\
    \Xhline{3\arrayrulewidth}\rule{0pt}{9pt}
    \end{tabular}
    }
    \vspace{-7mm}
    \caption{The results of ablation study for the effectiveness of PVP constraint and the conjunction of keyword categories. $\mathcal{O}$ indicates the result of only utilizing object-level keywords.}
    \vspace{-2mm}
    \label{table:ablation}
\end{table}

%% file: table/prompt_keyword.tex
%%%%%%%%%%%%%%%%%%%%%%%%%%%%%%%%%%%%%%%%%%%%%%%%%%%%%%%%%%%%%%%%%
\begin{table*}[t!]
\centering
\begin{minipage}{0.95\linewidth}\vspace{0mm}    
\centering
\begin{tcolorbox} 
\small
{\color[HTML]{3531FF} \textbf{Counterfactual Keywords Generation Prompt:}}

\#\#\#Instruction\#\#\# \\
Generate a list of counterfactual keywords for the provided image. These keywords should propose plausible yet intentionally misleading alternatives to the actual visual content of the image. Ensure that the changes are visually conceivable and logically consistent within the context of the scene. \\
\\
\#\#\#Guidelines\#\#\# \\
\textbf{(option. $\mathcal{O}$)} Object Substitution: Replace an object in the image with another that could logically occupy the same space but alters the scene's context or meaning.

\textbf{(option. $\mathcal{A}$)} Attribute Modification: Change an object’s color, size, or shape in a way that makes sense visually but leads to a different interpretation.

\textbf{(option. $\mathcal{R}$)} Relational Changes: Adjust the spatial or interactional relationships between objects to suggest a different narrative or dynamic within the scene. \\ \\
\#\#Examples\#\# \\
(Image 1): The photo features a tuxedo cat sitting inside the drum of a front-loading washing machine. The cat's distinctive white and black fur, white bib, and paws are visible against its dark body. It stares directly at the camera with bright eyes. The washing machine has various control knobs and buttons, and the area is cluttered with items like laundry detergent. The ambient, warm lighting adds a homely feel.

\textbf{(option. $\mathcal{O}$)}: small dog, laundry basket, robot vacuum, soccer ball

\textbf{(option. $\mathcal{A}$)}: orange cat, glowing dryer, vintage suitcase, oversized watch

\textbf{(option. $\mathcal{R}$)}: cat outside the dryer, dryer in a store display, cat playing with socks \\

\#\#\#Your Answer\#\#\# \\
List as many counterfactual keywords as possible for the image following the guidelines.

[Counterfactual Keywords]: 

\end{tcolorbox}
\vspace{-5mm}
\caption{Instruction prompt for generating counterfactual keywords. To generate different category of counterfactual keywords: object-, attribute-, or relation-level, the instruction has three options to choose $\mathcal{O}$, $\mathcal{A}$, or $\mathcal{R}$. }
\label{table:gen_prompt}
\end{minipage}
\vspace{-4mm}
\end{table*}
%%%%%%%%%%%%%%%%%%%%%%%%%%%%%%%%%%%%%%%%%%%%%%%%%%%%%%%%%%%%%%%%%

%% file: algo/algorithm.tex
% %%%%%%%%%%%%%%%%%%%%%%%%%%%%%%%%%%%%%%%%%%%%%%%%%%%%%%%%%%%%%%%%%%%%%%%%%%%%%%%%%%
% \begin{figure}[h!]
% \vspace{-0.4cm}
% \begin{algorithm}[H]
% \caption{Counterfactual Inception}
% \begin{algorithmic}[1]

% \Require visual inputs $v$, user query $q$, keyword generation prompt $p$, LMM $M_{\theta}$, CLIP distance \textbf{CLIP}.
% \State $M_{\theta}(v,p)$
% \While {$n<N$} \Comment{Keyword Generation}
% \State $k_{f}^{n}, k_{c}^{n} \gets \text{GPT-4V}(x_{v},x_{p})$ 
% \State set $n \gets n+1$
% \EndWhile

% \State $k_{c}^{v} \gets \{k_{c}^{v} \in k_{c} \mid \lambda_{\text{bot-K}\%} \leq \textbf{CLIP}(x_{v}, k_{c}) \leq \lambda_{\text{top-K}\%}\}$ \Comment{Visual Verification}
% \State $k_{c}^{*} \gets \{k_{c}^{l} \in k_{c}^{v} \mid \mathcal{R}_{\text{cont}}(k_{f}, k_{c}^{v}) \geq \tau \}$ \Comment{Linguistic Verification}

% %Dual-modality Verification Process

% % \State $k_{c}^{*} \gets$ DVP \Comment{Optimal Counterfactual Keywords}
% \State Model Response $\gets M_{\theta}(x_{v}, x_{i}, k_{c}^{*})$ \Comment{Counterfactual Inception}
% \end{algorithmic}
% \label{alg:1}
% \end{algorithm}
% \vspace{-0.4cm}
% \end{figure}
% %%%%%%%%%%%%%%%%%%%%%%%%%%%%%%%%%%%%%%%%%%%%%%%%%%%%%%%%%%%%%%%%%%

%%%%%%%%%%%%%%%%%%%%%%%%%%%%%%%%%%%%%%%%%%%%%%%%%%%%%%%%%%%%%%%%%%%%%%%%%%%%%%%%%%
\begin{figure}[h!]
\vspace{-0.55cm}
\begin{algorithm}[H]
\caption{Counterfactual Inception}
\begin{algorithmic}[1]
\Require Input image $v$, user query $q$, LMM $M_{\theta}$,  keyword generation prompt $p$ in Table.~\ref{table:gen_prompt}
% \Ensure Enhanced model response that mitigates hallucination
\State Initialize keyword lists $\mathcal{O}, \mathcal{A}, \mathcal{R}$
\For{$\text{c} \in \{\mathcal{O}, \mathcal{A}, \mathcal{R}\}$}  \Comment{Keyword gen \& PVP}
    \State $k \gets M_{\theta}\text{.generate}(v, p_{\text{c}})$
    \State $k_{\text{pvp}}\gets\{k{\in}\mathcal{|K|}{:}\lambda_{\text{bot}}{\leq}\textbf{CLIP}(v, k){\leq}\lambda_{\text{top}}\}$
    \State Append $k_{\text{pvp}}$ to category list.
\EndFor
\State $k^{*} \gets [\mathcal{O}; \mathcal{A}; \mathcal{R}]$ \Comment{Concatenate all keywords}

\While {$t<T$} \Comment{Implanting keywords}
    \State $\text{logit}_{M_{\theta}} \gets M_{\theta}(v, q, k, y_{<t})$
    \State $y_t = \text{argmax}(\text{Softmax}(\text{logit}_{M_{\theta}}))$
    \State Set $t \gets t+1$
\EndWhile

\State \Return $y_{<t+1}$ \Comment{Return generated responses}

\end{algorithmic}
\label{alg:counterfactual}
\end{algorithm}
\vspace{-0.7cm}
\end{figure}
%%%%%%%%%%%%%%%%%%%%%%%%%%%%%%%%%%%%%%%%%%%%%%%%%%%%%%%%%%%%%%%%%%

%% file: table/prompt_counterfactual.tex
%%%%%%%%%%%%%%%%%%%%%%%%%%%%%%%%%%%%%%%%%%%%%%%%%%%%%%%%%%%%%%%%%
\begin{table}[t!]
\centering
\begin{minipage}{0.95\linewidth}\vspace{0mm}    
\centering
\begin{tcolorbox} 
\small
{\color[HTML]{3531FF} \textbf{Counterfactual Prompt:}} 

Carefully avoid the listed Counterfactual Keywords in your response. \\

Counterfactual Keywords: \{{\color[HTML]{FF0000}\texttt{cf\_keywords}}\}.

Question: \{{\color[HTML]{FF0000}\texttt{question}}\}

\end{tcolorbox}
\vspace{-5mm}
\caption{Counterfactual prompt to integrate the generated counterfactual keywords with user queries. Note that \textcolor{red}{red} text indicates placeholders for the keywords and user questions.}
\label{table:cf_prompt}
\end{minipage}
\vspace{-5mm}
\end{table}
%%%%%%%%%%%%%%%%%%%%%%%%%%%%%%%%%%%%%%%%%%%%%%%%%%%%%%%%%%%%%%%%%

%% file: table/keyword_statistics.tex
\begin{table*}[t!]
    \centering
    \renewcommand{\tabcolsep}{2.0mm}
    \resizebox{1.0\linewidth}{!}{    
    \begin{tabular}{lccccccccccccccccc}
    \Xhline{3\arrayrulewidth}\rule{0pt}{9pt}
    \multirow{2}{*}{Model} & \multirow{2}{*}{PVP} & \multicolumn{4}{c}{POPE}   & \multicolumn{4}{c}{MMVP}   & \multicolumn{4}{c}{COCO}   & \multicolumn{4}{c}{MMHal-Bench} \\ \cmidrule(lr){3-6} \cmidrule(lr){7-10} \cmidrule(lr){11-14} \cmidrule(lr){15-18}
    \multicolumn{1}{c}{}                        &                   & $\mathcal{O}$    & $\mathcal{A}$ & $\mathcal{R}$ & Score     & $\mathcal{O}$  & $\mathcal{A}$ & $\mathcal{R}$ & Score     & $\mathcal{O}$  & $\mathcal{A}$ & $\mathcal{R}$ & Score     & $\mathcal{O}$   & $\mathcal{A}$ & $\mathcal{R}$ & Score       \\ \midrule
    \multirow{2}{*}{LLaVA-1.5}                  & \ding{55}                   & 571  & 557  & 678  & 0.205 & 457  & 487  & 568  & 0.199 & 796  & 858  & 907  & 0.204 & 142   & 159   & 124   & 0.201   \\
                                                & \Checkmark                   & 190  & 175  & 262  & 0.154 & 177  & 155  & 179  & 0.154 & 249  & 258  & 205  & 0.153 & 50    & 47    & 30    & 0.153   \\
    \multirow{2}{*}{IXC2-VL}                    & \ding{55}                   & 1963 & 1836 & 1783 & 0.189 & 1156 & 1087 & 1062 & 0.189 & 1928 & 1830 & 1838 & 0.188 & 371   & 352   & 344   & 0.191   \\
                                                & \Checkmark                   & 858  & 768  & 623  & 0.152 & 543  & 513  & 403  & 0.154 & 913  & 764  & 629  & 0.152 & 153   & 132   & 106   & 0.152   \\
    \multirow{2}{*}{LLaVA-NeXT}                 & \ding{55}                   & 2312 & 2120 & 1856 & 0.191 & 1333 & 1170 & 1092 & 0.192 & 2441 & 2172 & 2159 & 0.190 & 454   & 400   & 356   & 0.197   \\
                                                & \Checkmark                   & 1109 & 954  & 383  & 0.154 & 550  & 489  & 383  & 0.154 & 1070 & 897  & 781  & 0.153 & 180   & 159   & 140   & 0.154   \\
    \multirow{2}{*}{InternVL 1.5}               & \ding{55}                   & 1050 & 1039 & 1024 & 0.194 & 611  & 619  & 598  & 0.189 & 1034 & 1020 & 1071 & 0.191 & 203   & 197   & 192   & 0.193   \\
                                                & \Checkmark                   & 445  & 439  & 380  & 0.154 & 230  & 221  & 182  & 0.152 & 662  & 634  & 407  & 0.151 & 94    & 69    & 40    & 0.153   \\
    \multirow{2}{*}{Gemini 1.5}                 & \ding{55}                   & 1897 & 1795 & 1687 & 0.173 & 1191 & 1093 & 1090 & 0.178 & 1859 & 1832 & 1753 & 0.172 & 372   & 359   & 329   & 0.172   \\
                                                & \Checkmark                   & 1291 & 1028 & 582  & 0.151 & 749  & 630  & 377  & 0.151 & 1250 & 1090 & 632  & 0.150 & 230   & 184   & 108   & 0.150   \\
    \multirow{2}{*}{GPT4V}                      & \ding{55}                   & 2000 & 1922 & 1988 & 0.178 & 1200 & 1160 & 1182 & 0.181 & 1995 & 1865 & 1972 & 0.169 & 384   & 370   & 379   & 0.181   \\
                                                & \Checkmark                   & 1369 & 1021 & 549  & 0.153 & 748  & 656  & 314  & 0.151 & 1320 & 1211 & 732  & 0.150 & 234   & 184   & 90    & 0.150 \\ 
    \Xhline{3\arrayrulewidth}\rule{0pt}{9pt}
    \end{tabular}
    }
    \vspace{-8mm}
    \caption{Details of counterfactual keywords statistics and average CLIP score along keyword category.}
    \vspace{-3mm}
    \label{table:keyword}
\end{table*}